\documentclass[letterpaper, 10 pt, conference]{ieeeconf}  

\IEEEoverridecommandlockouts                              

\usepackage{amsmath} 
\usepackage{amssymb}  

\usepackage{url}
\usepackage{makecell}
\usepackage{multirow}
\usepackage{booktabs}
\usepackage[hidelinks]{hyperref}
\usepackage[table]{xcolor}
\usepackage{graphicx}

\newcommand{\figref}[1]{{Fig.~\ref{#1}}}
\newcommand{\bm}[1]{\mbox{\boldmath{$#1$}}}

\title{\LARGE \bf
  GuidedAttention: Interpretable and Correctable Visual Attention for OOD-Robust Robot Manipulation via Imitation Learning
}

\author{Masaki Murooka$^{1,2}$, Ryoichi Nakajo$^{2}$, Keisuke Shirai$^{2}$, Tomohiro Motoda$^{2}$, \\Hanbit Oh$^{2}$, Ryo Hanai$^{2}$, Yukiyasu Domae$^{2}$
  \thanks{This work was supported in part by JST CREST, Japan, Grant Number JPMJCR2553 and JSPS KAKENHI Grant Number 22K17984.}%
  \thanks{$^{1}$CNRS-AIST JRL (Joint Robotics Laboratory), IRL,
    1-1-1 Umezono, Tsukuba, Ibaraki 305-8560, Japan.
    {\tt\small m-murooka@aist.go.jp}}%
  \thanks{$^{2}$Artificial Intelligence Research Center,
    National Institute of Advanced Industrial Science and Technology (AIST),
    2-3-26 Aomi, Koto-ku, Tokyo 135-0064, Japan.}%
}

\begin{document}

\maketitle
\thispagestyle{empty}
\pagestyle{empty}

\setlength{\floatsep}{10pt}
\setlength{\textfloatsep}{10pt}
\setlength{\abovecaptionskip}{4pt}
\setlength{\abovedisplayskip}{6pt}
\setlength{\belowdisplayskip}{6pt}

\begin{abstract}
  End-to-end visuomotor policies provide little opportunity for humans to understand or correct the policy's visual attention. We propose GuidedAttention, a visuomotor imitation learning framework that introduces interpretable and correctable visual attention as an explicit intermediate representation. Task-relevant attention keypoints are predicted from camera images and condition a diffusion-based action policy. Users can inspect and optionally correct selected keypoints once at rollout initialization, after which the corrected attention is automatically propagated throughout execution by a tracking module. Experiments in simulation and the real world demonstrate that GuidedAttention consistently improves robot manipulation performance, particularly under positional and appearance out-of-distribution (OOD) conditions. \url{https://mmurooka.github.io/guided-attention-project-page}
\end{abstract}

\section{Introduction}
\label{sec:intro}

Learning robot behaviors directly from experience enables dexterous and
flexible manipulation even in environments that are difficult to model
explicitly using conventional model-based approaches.
End-to-end representation learning further eliminates the need for manual
feature engineering and can directly process high-dimensional sensory inputs.
However, compared with traditional modular pipelines, end-to-end policies
provide little opportunity for humans to understand or correct the policy's
visual attention during execution%
~\cite{MANIP:Yu:IROS2024,VideoControl:Zheng:arXiv2026}.
Even when a human can easily identify the task-relevant object or location,
the policy must discover such visual cues entirely on its own.
Consequently, a policy may fail simply because it lacks a simple but crucial
hint that a human could readily provide%
~\cite{RTTrajectory:Gu:ICLR2024,RoboticVisualInst:Li:CVPR2025}.

\begin{figure}[tpb]
  \centering
  \includegraphics[width=1.0\columnwidth]{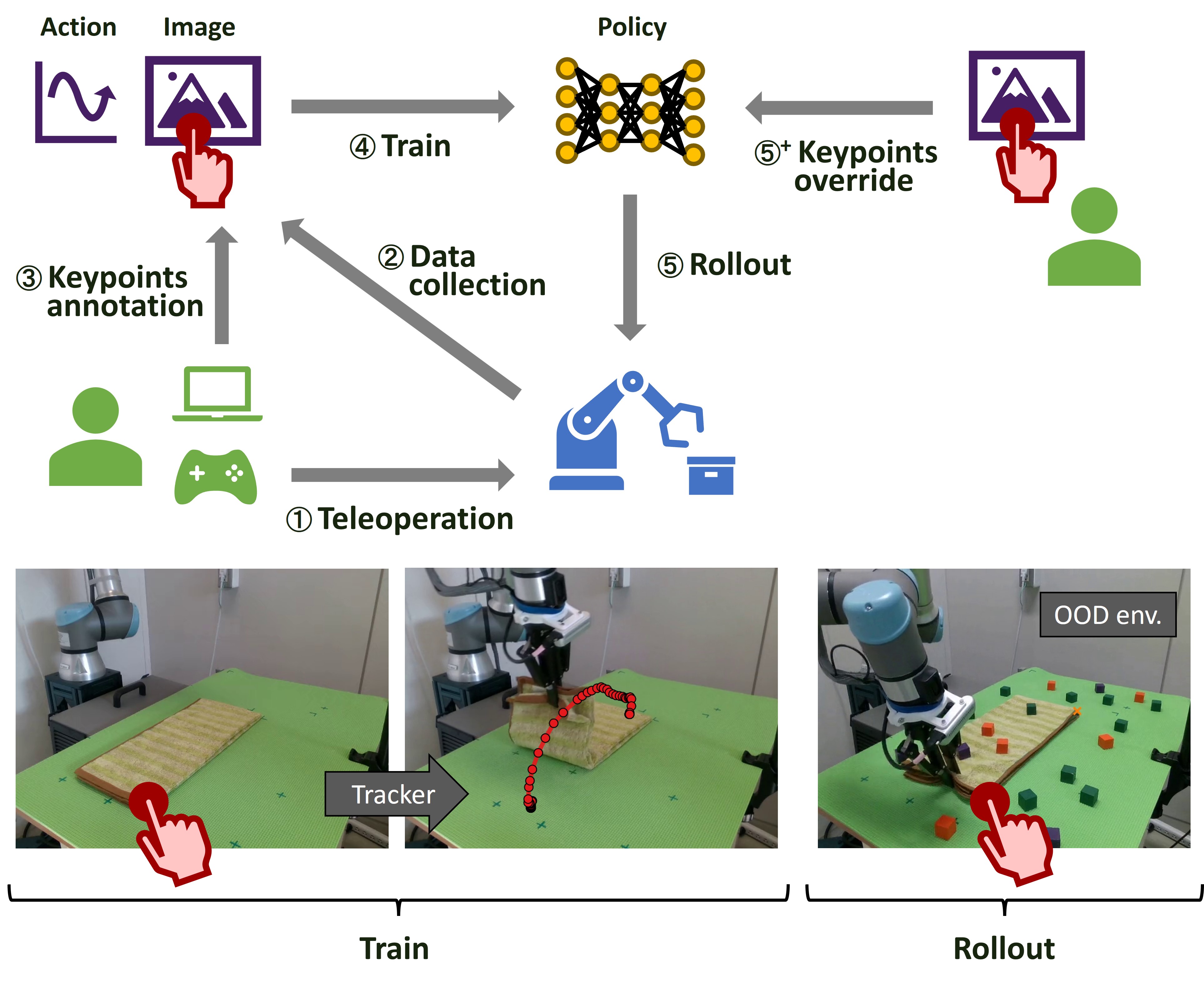}\\
  \caption{
    Overview of the GuidedAttention framework.
    \newline
    \footnotesize{
      Teleoperated demonstrations are annotated with attention keypoints only
      in the initial frame, which serve as an interpretable and correctable
      intermediate representation for visuomotor policy learning.
      During rollout, users may optionally correct selected keypoints once,
      after which they are automatically tracked throughout execution.
    }
  }
  \label{fig:intro}
\end{figure}

To address this limitation, we propose GuidedAttention, a visuomotor imitation
learning framework that introduces interpretable and correctable visual
attention as an explicit intermediate representation.
We instantiate this representation using task-relevant visual attention
keypoints predicted from camera images to condition a diffusion-based action
policy, termed Diffusion Policy with GuidedAttention (DP-GA).
For training, a user annotates the keypoints only in the initial frame of each
demonstration, and an off-the-shelf tracking model propagates them through the
remaining frames.
During rollout, the policy autonomously predicts both the attention keypoints
and robot actions.
When necessary, users may correct selected keypoints once at rollout
initialization, after which the corrected attention is automatically tracked
throughout execution (\figref{fig:intro}).
Thus, GuidedAttention preserves autonomous action generation while providing
an explicit interface through which users can inspect and correct the policy's
visual attention.

This capability is particularly useful in out-of-distribution (OOD)
environments, where visual attention learned from demonstrations may drift
away from task-relevant structures.
Imitation learning policies typically generalize well within the training
distribution but can suffer substantial performance degradation under such
distribution shifts%
~\cite{GenGap:Xie:ICRA2024,Analogy:Gupta:CoRL2025}.
GuidedAttention allows users to provide task-relevant visual guidance when
autonomous attention prediction becomes unreliable, without requiring
continuous supervision.
We evaluate the proposed method on multiple simulated and real-world
manipulation tasks and demonstrate improved performance under positional and
appearance OOD conditions.

The main contributions of this paper are summarized as follows:
\begin{itemize}
  \item We introduce interpretable and correctable visual attention as an
  explicit intermediate representation for visuomotor imitation learning,
  enabling users to inspect and optionally correct the policy's visual
  attention.

  \item We propose DP-GA, which conditions diffusion-based action generation
  on visual attention keypoints and supports one-time user correction followed
  by automatic tracking.

  \item We demonstrate through simulation and real-world experiments that
  GuidedAttention improves robot manipulation performance, particularly under
  positional and appearance OOD conditions.
\end{itemize}

\section{Related Works} \label{sec:related-works}

\subsection{Visual Attention in Imitation Learning}

Research on visual imitation learning increasingly focuses on extracting
task-relevant information from visual observations to improve policy learning.
Early work demonstrated that representing observations through visual
keypoints enables more structured visuomotor control%
~\cite{SpatialAutoencoders:Finn:ICRA2016}.

Recent approaches have explored how such attention cues can be obtained.
Attention can be estimated by predicting keypoints that minimize visual
dynamics prediction errors~\cite{SARNN:Ichiwara:ICRA2022},
derived from object proposal priors~\cite{VIOLA:Zhu:ICLR2023},
or extracted from human demonstration videos%
~\cite{LearningByWatching:Xiong:IROS2021}.
Another line of work investigates how these representations should be utilized
within visuomotor policies.
Keypoints can serve as the primary visual representation%
~\cite{KVIL:Gao:TRO2022},
be enriched with semantic object-part information%
~\cite{SKIL:Wang:RSS2025},
or be selectively filtered to improve robustness%
~\cite{ATK:Zhang:CoRL2025}.

Unlike previous methods, which primarily use attention as a learned
representation for improving policy performance, GuidedAttention treats visual
attention as an explicit intermediate representation that is both
interpretable and correctable by users.
This attention representation provides an intuitive interface between human
guidance and end-to-end visuomotor policies while remaining tightly coupled
with action generation.




\subsection{Human Guidance for Visuomotor Policies}

Human guidance for visuomotor policies can be provided at different levels of
abstraction.
Language-based and vision-language models specify high-level task goals%
~\cite{SayCan:Brohan:CoRL2022,VIMA:Jiang:ICML2023},
whereas visual prompting methods provide more direct spatial guidance through
bounding boxes%
~\cite{VPrompt:Muttaqien:CASE2025},
sketches~\cite{RTSketch:Sundaresan:CoRL2024},
trajectory annotations%
~\cite{RTTrajectory:Gu:ICLR2024},
semantic annotations%
~\cite{RoboticVisualInst:Li:CVPR2025},
or sparse point prompts~\cite{P3PO:Levy:ICRA2025}.
While these methods differ in how human guidance is provided, they
all demonstrate the effectiveness of incorporating human-provided information
to improve visuomotor policies.

GuidedAttention takes a different approach by introducing an explicit
intermediate attention representation between perception and action generation.
Instead of directly specifying guidance to the policy, users interact with the
policy through its predicted attention keypoints, which can be inspected and
optionally corrected before action generation.





\subsection{OOD Robustness in Imitation Learning}

OOD issues in imitation learning arise in various forms, including positional
OOD where object configurations differ from training data, appearance OOD
caused by changes in background or texture, and category OOD involving unseen
objects.
Such distribution shifts often cause the policy to lose focus on task-relevant
entities, resulting in degraded manipulation performance.

Existing approaches largely improve robustness in two ways.
One direction explicitly expands the training distribution by incorporating
OOD states into the data.
BMIL generates additional trajectories around demonstrations%
~\cite{BMIL:Park:NeurIPS2022},
RoCoDA counterfactually replaces distracting environmental elements%
~\cite{RoCoDA:Ameperosa:ICRA2025},
and ROSIE synthesizes semantic visual variations%
~\cite{ROSIE:Yu:RSS2023}.
Although these methods increase data diversity, their robustness remains
limited to the variations encountered during training.

A second direction focuses on learning representations that remain robust under
distribution shifts.
Latent Policy Barrier constrains policy rollouts within the learned
in-distribution latent space%
~\cite{LPBarrier:Sun:arXiv2025},
while Adapting by Analogy transfers manipulation strategies through functional
correspondence across object categories%
~\cite{Analogy:Gupta:CoRL2025}.

Unlike these approaches, GuidedAttention provides an explicit mechanism for
correcting task-relevant visual attention when autonomous prediction becomes
unreliable.
This complementary human-guided correction enables the policy to recover
task-relevant visual focus under positional and appearance OOD conditions.

\section{Method} \label{sec:method}

\begin{figure}[tpb]
  \centering
  \includegraphics[width=0.90\columnwidth]{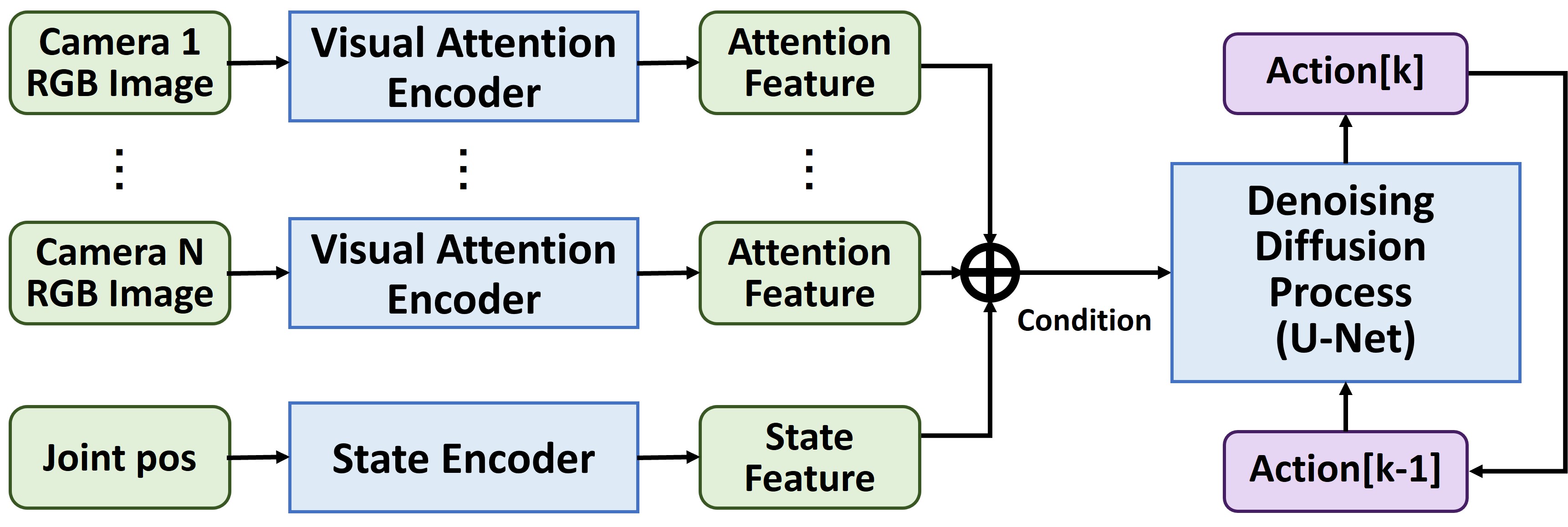}\\
  \vspace{1.5mm}
  \footnotesize (A) Overall policy architecture\\
  \vspace{4mm}
  \includegraphics[width=0.97\columnwidth]{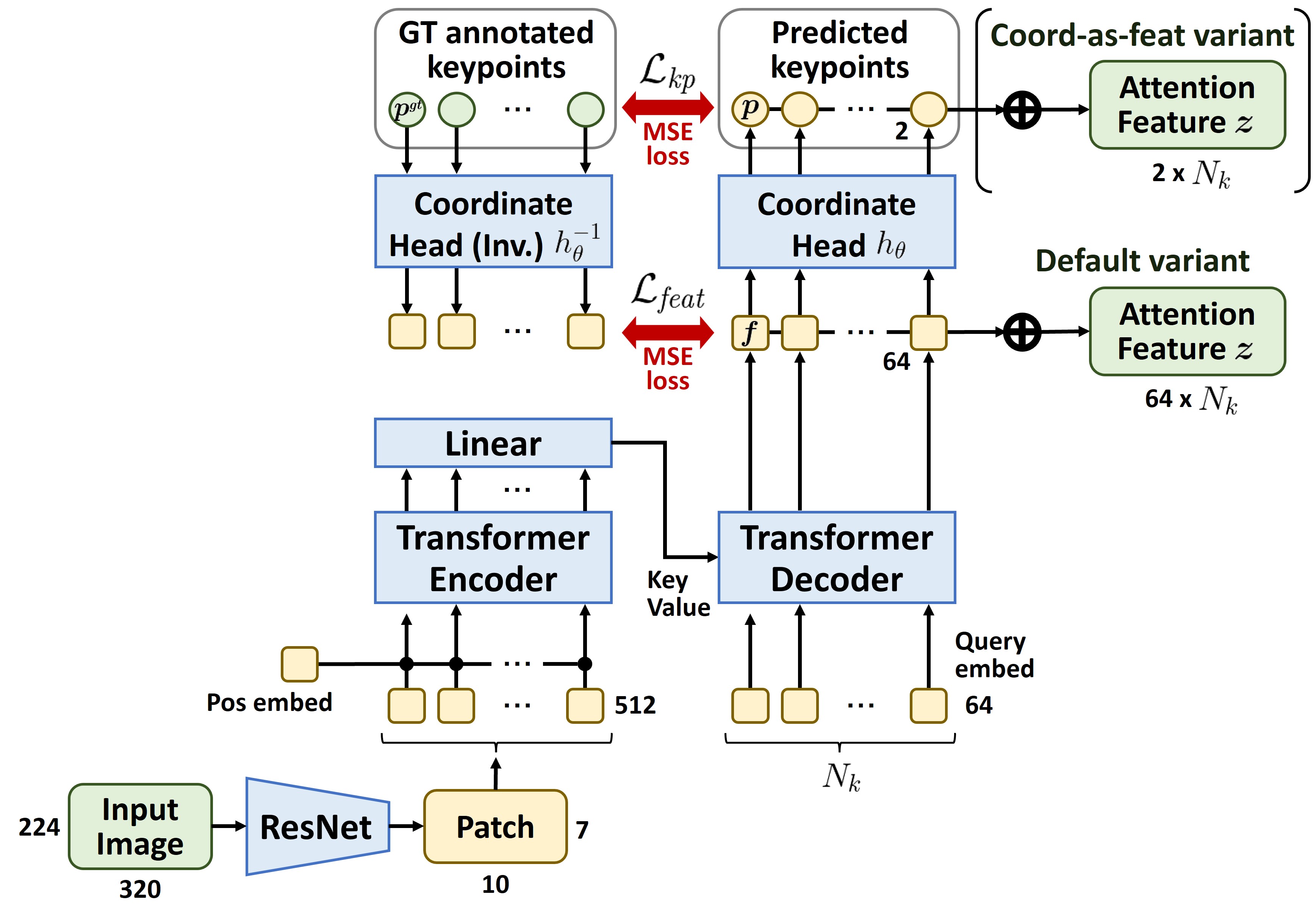}\\
  \footnotesize (B) Visual attention encoder\\
  \vspace{4mm}
  \includegraphics[width=0.655\columnwidth]{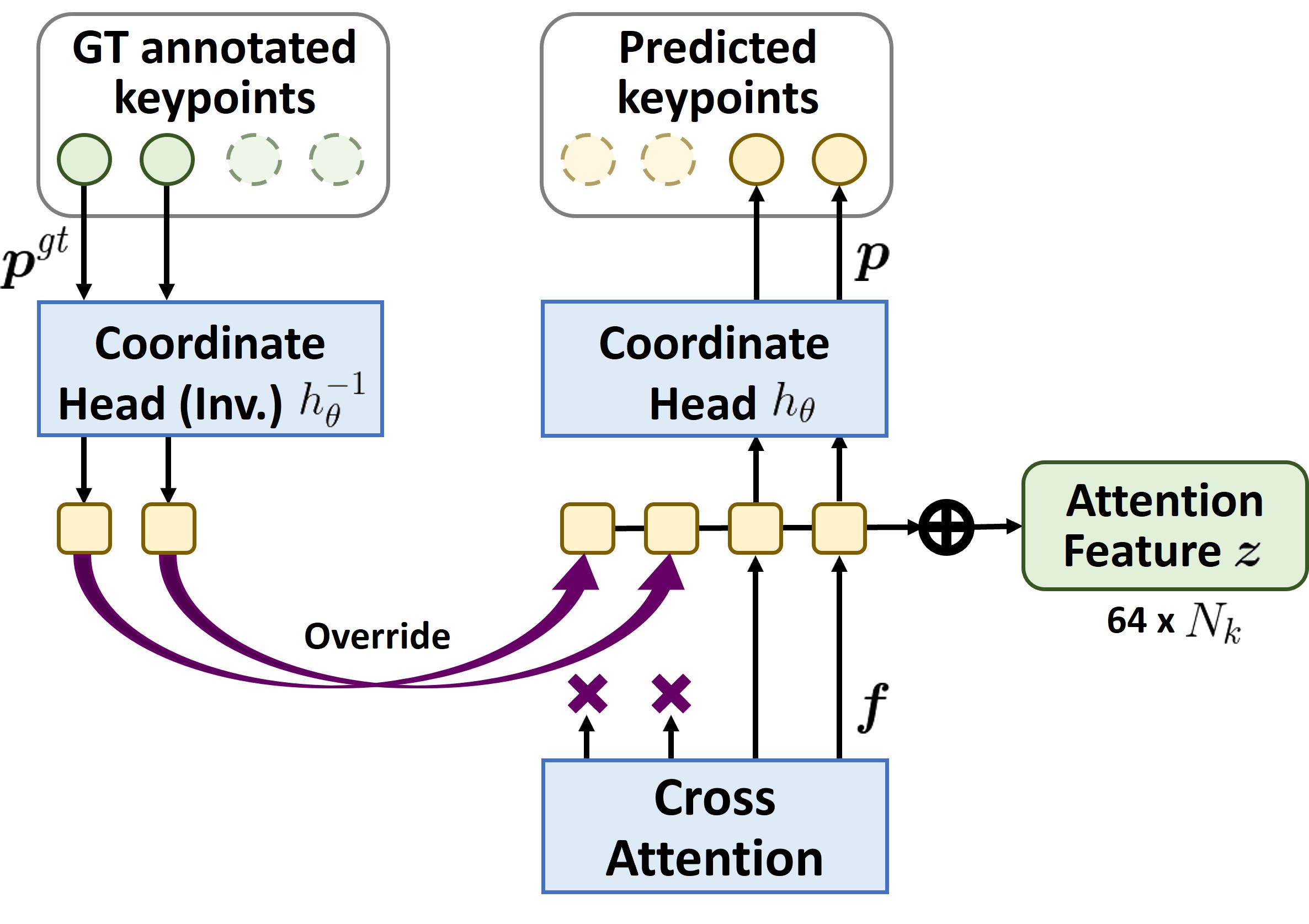}\\
  \footnotesize (C) Keypoint override mechanism
  \caption{
    Policy architecture overview.
    \newline
    \footnotesize{
      (A) The policy conditions a denoising diffusion process for action generation on attention features from image-based keypoints, together with proprioceptive states.\newline
      (B) Each image is processed by a ResNet and Transformer encoder-decoder to predict pixel-level keypoints, which are embedded as attention features.\newline
      (C) During rollout, users can optionally correct predicted keypoints when attention drift occurs. The corrected keypoints replace the predictions and update the attention features, enabling recovery from failures. The figure shows two of four keypoints overridden.
  }}
  \label{fig:policy}
\end{figure}

\subsection{Overall Policy Architecture}

An overview of the proposed DP-GA is shown in \figref{fig:policy}~(A).
Unlike conventional visuomotor policies that directly condition action
generation on learned visual features, DP-GA introduces an explicit visual
attention representation between perception and action generation.
This attention representation is interpretable by users and can be optionally
corrected during policy rollout while remaining tightly coupled with action
generation.

DP-GA takes multi-view RGB observations and proprioceptive joint states as
input.
For each camera image, a visual attention encoder (AttnEnc) predicts
task-relevant attention keypoints and extracts corresponding attention
features.
These attention features are concatenated with proprioceptive features and
used to condition a CNN-based denoising diffusion policy built on a U-Net
backbone~\cite{DP:Chi:IJRR2025}.

Unlike prior diffusion policies conditioned on global visual features from
CNNs~\cite{DP:Chi:IJRR2025} or point cloud encoders~\cite{DP3D:Ze:RSS2024},
DP-GA conditions action generation on interpretable visual attention features
that explicitly localize task-relevant structures while providing an explicit
interface for optional human correction.


\subsection{Visual Attention Encoder}

The AttnEnc, shown in \figref{fig:policy}~(B), predicts the explicit
intermediate representation of visual attention.
It estimates task-relevant attention keypoints together with their
corresponding attention features, which are subsequently used to condition
the diffusion policy.
AttnEnc follows a DETR-style architecture~\cite{DETR:Carion:ECCV2020}
composed of a Transformer encoder and decoder, but instead of predicting
bounding boxes as in DETR, it predicts 2D coordinates of attention
keypoints.
RGB images are first processed by a ResNet-18 backbone~\cite{ResNet:He:CVPR2016} to obtain patch features, which are input to the Transformer encoder with positional embeddings.
The decoder receives learnable queries corresponding to a pre-defined number of keypoints and produces attention features.
Finally, a coordinate head implemented as a linear layer maps each attention feature to a 2D coordinate as:
\begin{align}
\bm{p}_i = h_\theta(\bm{f}_i) = \bm{A}\bm{f}_i + \bm{b}
\end{align}
where $\bm{p}_i \in \mathbb{R}^2$ denotes the coordinate of the $i$-th keypoint and
$\bm{f}_i \in \mathbb{R}^{d_{\!f}}$ is its attention feature.
The parameters of the linear mapping $h_\theta(\cdot)$ are represented by $\bm{A}$ and $\bm{b}$.
For each image, AttnEnc outputs a set of attention features
$\{\bm{f}_i\}_{i=1}^{N_k}$, where $N_k$ is the pre-defined number of
task-relevant keypoints. These attention features are then used to condition
the denoising diffusion process for action generation.

We also evaluate a coordinate-as-feature variant, in which the predicted
coordinates $\bm{p}_i$ themselves, instead of $\bm{f}_i$, are used
as the conditioning inputs to the diffusion process.

Training optimizes both action prediction and keypoint prediction.
Let $\bm{p}^{\mathit{gt}}_i \in \mathbb{R}^2$ denote the ground-truth
coordinate of the $i$-th keypoint obtained from human annotation. The
keypoint loss is defined as the mean squared error (MSE) between predicted
and ground-truth keypoints:
\begin{align}
\mathcal{L}_{kp} =
\frac{1}{N_k}\sum_{i=1}^{N_k}
\lVert \bm{p}_i - \bm{p}^{\mathit{gt}}_i\rVert_2^2
\end{align}

By conditioning the diffusion process on features supervised through keypoint prediction,
the policy learns to focus on spatially relevant locations that are essential for manipulation.
At rollout, actions are generated by predicting keypoints from the current images via AttnEnc and conditioning the diffusion model on the resulting attention features.




\subsection{Keypoint Override Mechanism}

During policy rollout, the predicted attention keypoints serve as an
interpretable interface through which users can inspect the policy's visual
attention.
When the predicted attention is judged to be unreliable (e.g., under
appearance or positional OOD conditions), users may optionally correct a
subset of the keypoint coordinates, as illustrated in
\figref{fig:policy}~(C).

The corrected coordinates are then converted into attention features via
the inverse mapping of $h_\theta(\cdot)$ and replace the corresponding
predicted features in the final conditioning set:
\begin{align}
\bm{z}_i =
\begin{cases}
\bm{f}_i & \text{(prediction path)}\\[4pt]
h_\theta^{-1}(\bm{p}^{\mathit{gt}}_i) & \text{(override path)}
\end{cases}
\label{eq:keypoint-override}
\end{align}
where $\bm{z}_i$ denotes the feature actually used for policy conditioning.

Three techniques are introduced to ensure that this override mechanism is consistent and reliable:

(i) \textbf{Forward--inverse weight sharing.}
Since $h_\theta(\cdot)$ is a linear mapping from a $d_{\!f}$-dimensional feature to a 2D coordinate,
its analytical inverse can be defined without additional parameters:
\begin{align}
h_\theta^{-1}(\bm{p}_i)
= \bm{A}^{\!+}(\bm{p}_i - \bm{b})
\end{align}
where $\bm{A}^{\!+}$ denotes the Moore-Penrose pseudoinverse.
Enforcing this parameter sharing encourages mutual consistency between forward prediction and inverse recovery.

(ii) \textbf{Feature-space alignment loss.}
Because the inverse mapping is underdetermined when $d_{\!f} > 2$,
multiple feature vectors may correspond to the same coordinate.
To ensure that forward-predicted and inverse-restored features remain aligned,
we introduce the following feature-space consistency loss:
\begin{align}
\mathcal{L}_{\mathit{feat}}
= \frac{1}{N_k}\sum_{i=1}^{N_k}
\left\lVert
\bm{f}_i - h_\theta^{-1}(\bm{p}^{\mathit{gt}}_i)
\right\rVert_2^2
\end{align}
This encourages both paths to produce consistent keypoint features.

(iii) \textbf{Randomized feature routing.}
Even after applying the above alignment strategy, a residual
discrepancy may remain between forward-predicted features and those restored
from ground-truth keypoints. To avoid the diffusion model depending on one
specific pathway during training, we randomly select whether each keypoint feature $\bm{z}_i$
comes from the prediction path or the override path when conditioning the
diffusion process. This encourages both feature types to induce consistent
action distributions and stabilizes early training when predicted keypoints
are still inaccurate.

\subsection{Keypoint Annotation and Tracking}

\subsubsection{Training: ground-truth label generation}
To obtain ground-truth keypoints for supervision, a human clicks the
keypoint locations on the first frame of each demonstration, which
keeps the annotation lightweight and intuitive. The subsequent
ground-truth keypoints are then obtained by tracking:
\begin{align}
\bm{p}^{\mathit{gt}}_i[t]
= G(\bm{p}^{\mathit{gt}}_i[t-1],\ I[t],\ I[t-1]),
\end{align}
where $G(\cdot)$ is a tracking module (Co-Tracker~\cite{CoTracker:Karaev:ECCV2024} in our implementation),
and $I[t]$ is the RGB image at time $t$. Keypoints that are known to be
static remain fixed throughout the trajectory.


\subsubsection{Rollout: human-controllable attention}
At test time, the policy predicts keypoints in a feedforward manner,
enabling fully autonomous execution without human guidance. Since the
predicted keypoints serve as a human-interpretable intermediate
representation, the policy's visual focus can be monitored in real time
and failure modes can be diagnosed.

To enhance robustness under challenging visual conditions such as OOD
conditions, the user may optionally provide corrected keypoints at the
initial rollout frame by clicking on camera images. The override can be
applied to only a subset of keypoints or to selected camera views, reducing
interaction cost. The tracking module then continuously propagates these
corrected keypoints and supplies them as conditioning inputs to the
diffusion process throughout execution, while static keypoints remain
fixed. This enables human-controllable attention with minimal effort.




\section{Experiments} \label{sec:exp}

\subsection{Simulation Experiments}

\subsubsection{Tasks and Setup}

\begin{figure}[tpb]
  \centering
  \includegraphics[width=0.45\columnwidth]{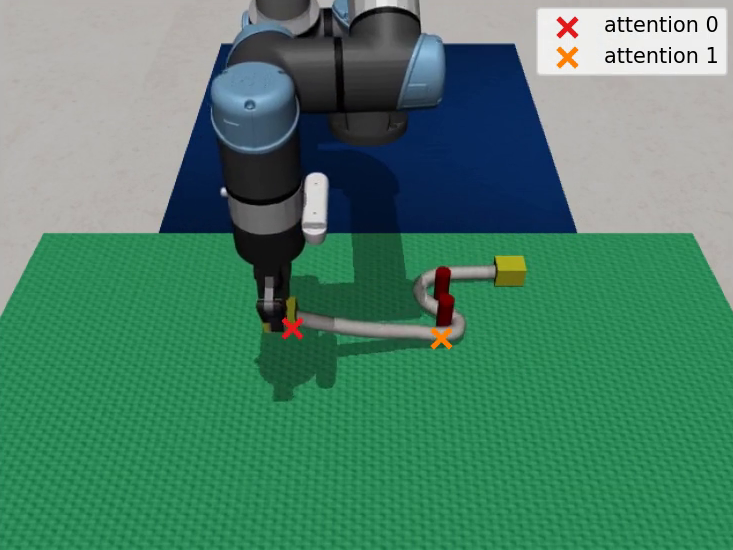}
  \includegraphics[width=0.45\columnwidth]{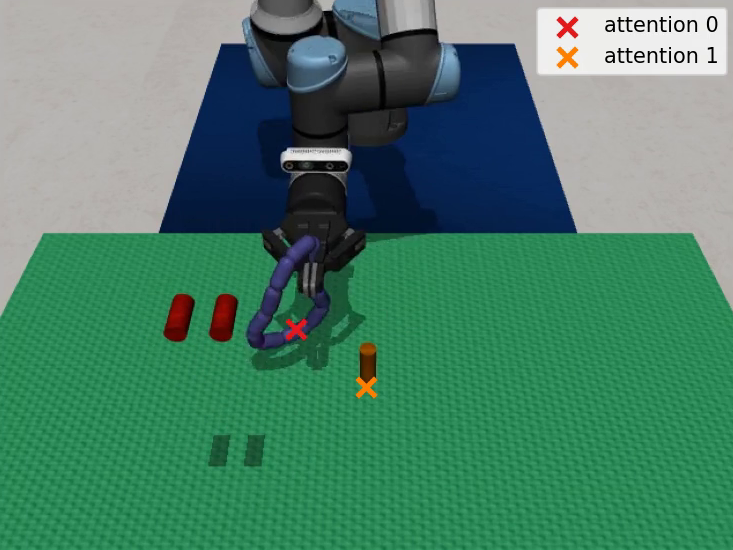}\\
  \begin{minipage}{0.45\columnwidth}
    \begin{center} \footnotesize (A) Cable \end{center}
  \end{minipage}
  \begin{minipage}{0.45\columnwidth}
    \begin{center} \footnotesize (B) Ring \end{center}
  \end{minipage}\\
  \vspace{2mm}
  \includegraphics[width=0.45\columnwidth]{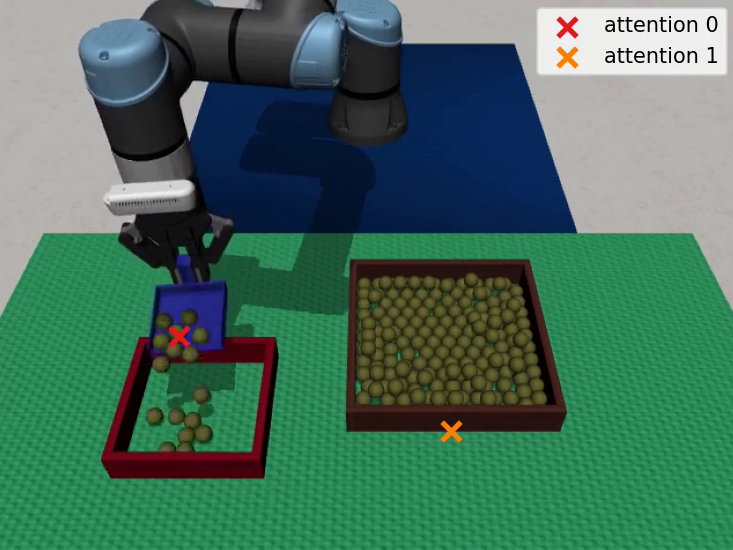}
  \includegraphics[width=0.45\columnwidth]{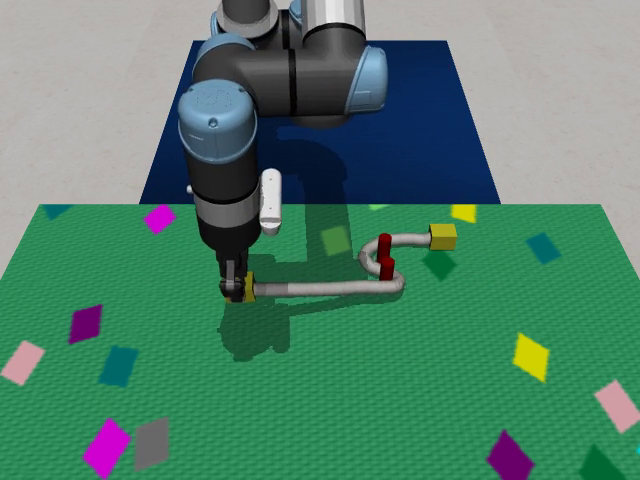}\\
  \begin{minipage}{0.45\columnwidth}
    \begin{center} \footnotesize (C) Particle \end{center}
  \end{minipage}
  \begin{minipage}{0.45\columnwidth}
    \begin{center} \footnotesize (D) Cable (Appearance OOD) \end{center}
  \end{minipage}
  \caption{
    Simulation manipulation tasks.
    \newline
    \footnotesize{
      (A) Cable: passing a flexible cable through a narrow gap between poles.
      (B) Ring: placing a flexible ring around a pole.
      (C) Particle: scooping particles into a box.
      (D) Cable under Appearance OOD: textured distractors on the tabletop introduce severe visual changes.
      All tasks use two attention keypoints: a dynamic keypoint attached to the manipulated object or tool,
      and a static keypoint defined on the target object or environment.
  }}
  \label{fig:sim-tasks}
\end{figure}

We evaluate the proposed method in the MuJoCo simulation environment using
three manipulation tasks that require dexterous and precise visuomotor coordination (\figref{fig:sim-tasks}).
All tasks involve a Universal Robots UR5e robotic arm equipped with a parallel gripper and a
single RGB camera providing a fixed diagonal overhead view.

\textbf{Cable}:
The robot grasps the end of a flexible cable and guides it through a narrow
gap between two vertical poles on the table.
We use two attention keypoints: a dynamic keypoint on the grasped cable end
and a static keypoint at the base of one pole.

\textbf{Ring}:
The robot grasps a ring and places it over a standing pole such that the ring
encircles the pole and rests on the table surface.
We use two attention keypoints: a dynamic keypoint on the lowest part of the
ring at the start and a static keypoint at the base of the pole.

\textbf{Particle}:
The robot uses a scoop-like tool to transfer particles from a source box to a
target box.
We use two attention keypoints: a dynamic keypoint on the scoop tip and a
static keypoint at the edge of the source box.

For each task, we collect a total of 30 teleoperated demonstrations
(15 per each of two in-distribution target object positions).
Ground-truth attention keypoints are manually annotated at the first frame of
each demonstration. Dynamic keypoints are then automatically propagated using
Co-Tracker~\cite{CoTracker:Karaev:ECCV2024}, while static keypoints remain fixed.

All experiments, in both simulation and the real world, are implemented using
RoboManipBaselines framework~\cite{RoboManipBaselines:Murooka:2025}.

\subsubsection{Evaluation Conditions}

We evaluate performance under both in-distribution (ID) and OOD conditions.

\textbf{Positional OOD}:
In ID conditions, rollouts are executed at 10 target-object positions
interpolated between the two target positions used in the demonstrations.
In OOD conditions, rollouts are performed at 10 unseen target-object positions
that either extrapolate beyond the demonstrated range or shift perpendicularly
to the demonstrated variation axis.
The target object corresponds to the poles in Cable and Ring, and the source
box in Particle.

\textbf{Appearance OOD}:
In ID conditions, the tabletop has a solid green texture.
In OOD conditions, we augment the same green tabletop with a cluttered pattern
of colorful square patches that are randomly placed and oriented, introducing
significant appearance changes.

\subsubsection{Comparison with Baselines}

\begin{table*}[tpb]
\centering
\caption{Simulation success rates across four generalization conditions}
\label{tab:sim-results}
\begin{tabular}{lllcccc}
\toprule
Policy & Variant & Correction & ID & Pos-OOD & App-OOD & Pos+App-OOD \\
\midrule
\rowcolor{green!10}DP-GA & Default & - & \textbf{83.3\% {\tiny ($\pm$7.4\%)}} & 55.6\% {\tiny ($\pm$8.6\%)} & 45.6\% {\tiny ($\pm$8.8\%)} & 34.4\% {\tiny ($\pm$6.9\%)} \\
\rowcolor{red!10}DP-GA & Default & Enabled & \textbf{84.4\% {\tiny ($\pm$4.3\%)}} & \textbf{70.0\% {\tiny ($\pm$5.5\%)}} & \textbf{76.7\% {\tiny ($\pm$13.1\%)}} & \textbf{67.8\% {\tiny ($\pm$13.7\%)}} \\
\rowcolor{green!10}DP-GA & Coord-as-feat & - & \textbf{84.4\% {\tiny ($\pm$4.3\%)}} & 55.6\% {\tiny ($\pm$13.5\%)} & 70.0\% {\tiny ($\pm$5.7\%)} & 38.9\% {\tiny ($\pm$13.6\%)} \\
\rowcolor{red!10}DP-GA & Coord-as-feat & Enabled & \textbf{87.8\% {\tiny ($\pm$5.7\%)}} & \textbf{68.9\% {\tiny ($\pm$6.9\%)}} & \textbf{81.1\% {\tiny ($\pm$10.6\%)}} & \textbf{63.3\% {\tiny ($\pm$7.2\%)}} \\
\midrule
DP & - & - & 67.8\% {\tiny ($\pm$6.3\%)} & 45.6\% {\tiny ($\pm$9.6\%)} & 54.4\% {\tiny ($\pm$9.0\%)} & 28.9\% {\tiny ($\pm$5.7\%)} \\
ACT & - & - & 53.3\% {\tiny ($\pm$30.7\%)} & 28.9\% {\tiny ($\pm$14.1\%)} & 6.7\% {\tiny ($\pm$5.7\%)} & 8.9\% {\tiny ($\pm$6.9\%)} \\
\bottomrule
\end{tabular}\\
\vspace{2mm}
\begin{minipage}{1.8\columnwidth}
\footnotesize{
Results averaged over three manipulation tasks (Cable, Ring, Particle).
Each value indicates mean success rate $\pm$ standard deviation across
3 random seeds and 3 tasks, each with 10 rollouts for every ID / OOD condition.
Green rows: fully autonomous execution.
Red rows: optional human correction of the predicted attention at the first rollout frame, followed by automatic tracking.
Bold numbers denote the best performance and those within 5 percentage points of the best in each condition.
}
\end{minipage}
\end{table*}

We compare our approach against two strong vision-based imitation learning baselines.
We evaluate two variants of our guided-attention policy.

\textbf{DP-GA (default)}:
The diffusion model is conditioned on attention features extracted from a
visual attention encoder, where keypoint prediction encourages their spatial
alignment to task-relevant physical structures.

\textbf{DP-GA (coordinate-as-feature)}:
The diffusion model is directly conditioned on the predicted attention
keypoint coordinates, treating them as structured geometric cues for control.

Both variants share the same visual attention encoder and the guided-attention
override mechanism described in Sec.~\ref{sec:method}, and differ only in how
the attention representation conditions the policy.

For comparison, we include two widely used vision-based imitation learning baselines.

\textbf{DP}:
Diffusion policy conditioned on CNN-based global visual features
without explicit attention representation~\cite{DP:Chi:IJRR2025}.

\textbf{ACT}:
An action-chunking transformer that predicts multi-step action sequences
from visual observations~\cite{ACT:Zhao:RSS2023}.

All methods are trained and evaluated with identical datasets and rollout
protocols: 10 rollouts per condition for each of 3 random seeds.

\textbf{Performance without human intervention.}
When executed fully autonomously, DP-GA corresponds to the green rows in
Table~\ref{tab:sim-results}.
Comparing the baselines first, ACT performs similarly to DP under ID
conditions, but suffers a large performance drop in OOD settings, indicating
the strong generalization ability of diffusion models.
On top of that, DP-GA further improves over DP across most evaluation
conditions.
In particular, it achieves approximately a 15 percentage points (pp) improvement under ID,
demonstrating the benefit of leveraging spatially structured attention cues.
Although DP-GA (default) slightly underperforms DP in the appearance OOD
setting, DP-GA consistently outperforms DP by around 10\,pp in all other OOD
conditions, confirming that guided-attention learning significantly enhances
robust visuomotor control under distribution shift.

\textbf{Additional gains from attention correction.}
When the predicted attention becomes unreliable, users may optionally
correct the attention keypoints only at the first frame, after which
they are automatically tracked throughout execution.
This leads to substantial performance improvements under distribution
shift.
While maintaining a high success rate in ID conditions,
both DP-GA variants improve OOD performance by approximately
15--30\,pp over their fully autonomous execution.
Consequently, DP-GA achieves around 20--30\,pp higher success rates
than the DP baseline in both positional and appearance OOD settings
(see Table~\ref{tab:sim-results}, red rows).
These results indicate that optional human correction effectively
complements autonomous attention prediction and further improves
performance under positional and appearance distribution shifts.

\subsubsection{Ablation Studies}

\begin{table}[tpb]
\centering
\caption{Ablation of guided-attention components}
\label{tab:ablation-sim}
\begin{tabular}{llc}
\toprule
Variant & Correction & ID ($\Delta$ from Default) \\
\midrule
\rowcolor{green!10}No feature routing & - & 81.1\% {\tiny ($\pm$6.9\%)} (-2.2 pp) \\
\rowcolor{red!10}No feature routing & Enabled & 25.5\% {\tiny ($\pm$14.6\%)} (\textbf{-58.9 pp}) \\
\rowcolor{green!10}No attention supervision & - & 52.2\% {\tiny ($\pm$7.3\%)} (\textbf{-31.1 pp}) \\
\rowcolor{red!10}No attention supervision & Enabled & 17.8\% {\tiny ($\pm$16.7\%)} (\textbf{-66.7 pp}) \\
\bottomrule
\end{tabular}\\
\vspace{2mm}
\begin{minipage}{0.93\columnwidth}
\footnotesize{
Ablation applied to the default variant of DP-GA.
All results are obtained under the same evaluation protocol as Table~\ref{tab:sim-results}.
Values indicate mean success rate $\pm$ standard deviation across 3 simulation tasks
(Cable, Ring, Particle), 3 random seeds, and 10 rollouts per ID condition.
The numbers in parentheses show performance drops from DP-GA (default).
}
\end{minipage}
\end{table}

To assess the contribution of the key guided-attention components, we conduct
ablation studies on the default variant of DP-GA.

\textbf{Feature routing}:
During training, randomized routing between the prediction path and the override
path is disabled. The diffusion model is therefore conditioned solely on
predicted attention features.

\textbf{Attention supervision}:
The keypoint loss $\mathcal{L}_{\mathit{kp}}$ and the feature-alignment loss
$\mathcal{L}_{\mathit{feat}}$ are removed during training, so that attention features are
not enforced to represent spatially meaningful locations.

\textbf{Ablation analysis.}
Table~\ref{tab:ablation-sim} summarizes the performance degradation when ablating key guided-attention components from the default variant of DP-GA, evaluated under ID conditions with and without attention correction.
Disabling feature routing causes a drastic drop of nearly 60\,pp when attention correction is enabled, indicating that proper integration of override-based features during training is critical for leveraging human attention correction during rollout.
In contrast, removing attention supervision leads to significant degradation both with and without attention correction (approximately 65\,pp and 30\,pp drops, respectively), showing that robustness requires attention features to be aligned with task-relevant structures.
These results confirm that robustness does not simply arise from the transformer-feature-conditioned diffusion policy architecture itself; rather, it critically relies on correctly integrating human-corrected attention into policy conditioning.
Overall, our simulation results demonstrate that guided attention substantially improves robustness to positional and appearance distribution shifts.

\subsubsection{Additional Analysis}

We further analyze the proposed method from the perspectives of robustness and prompting.

\begin{figure}[tpb]
  \centering
  \includegraphics[width=1.0\columnwidth]{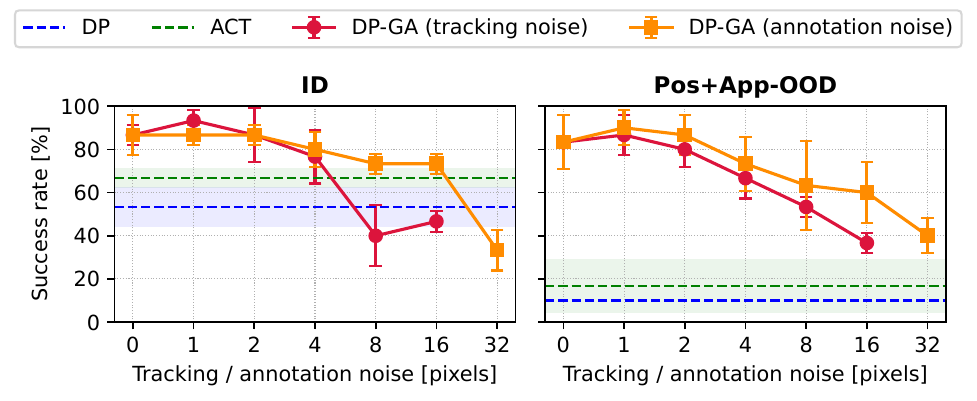}\\
  \vspace{-2mm}
  \caption{Quantitative impact of tracking and annotation accuracy.}
  \label{fig:eval-noise}
\end{figure}

\textbf{Robustness to tracking and annotation accuracy.}
\figref{fig:eval-noise} shows the impact of noise added to tracked and user-specified keypoints.
The method remains stable under moderate noise and degrades gradually, while maintaining clear advantages over baseline policies under OOD conditions.

\textbf{Comparison with marker overlay prompting.}
Table~\ref{tab:eval-vp} compares with baselines that provide visual prompts by directly overlaying bounding boxes or point markers on the input images~\cite{VPrompt:Muttaqien:CASE2025}.
While these approaches provide similar spatial cues, they tend to be less effective, particularly under OOD conditions.

\begin{table}[t]
\centering
\caption{Comparison with marker overlay prompting}
\label{tab:eval-vp}
\begin{tabular}{lll}
\toprule
Policy & ID & Pos+App-OOD \\
\midrule
DP-GA & \textbf{86.7\% {\tiny ($\pm$4.7\%)}} & \textbf{83.3\% {\tiny ($\pm$12.5\%)}} \\
MOP-Box & 60.0\% {\tiny ($\pm$0.0\%)} & 33.3\% {\tiny ($\pm$24.9\%)} \\
MOP-Point & 50.0\% {\tiny ($\pm$14.1\%)} & 3.3\% {\tiny ($\pm$4.7\%)} \\
\bottomrule
\end{tabular}
\end{table}

\textbf{Impact of the number of keypoints and cameras.}
Strong performance is achieved without increasing the number of keypoints or camera views, as long as task-relevant ones are selected.
Both keypoints and cameras can be partially overridden, enabling flexible human correction without requiring full
re-specification.

\subsection{Real-World Experiments}

\subsubsection{Tasks and Setup}

\begin{figure*}[tpb]
  \centering
  \includegraphics[width=0.39\columnwidth]{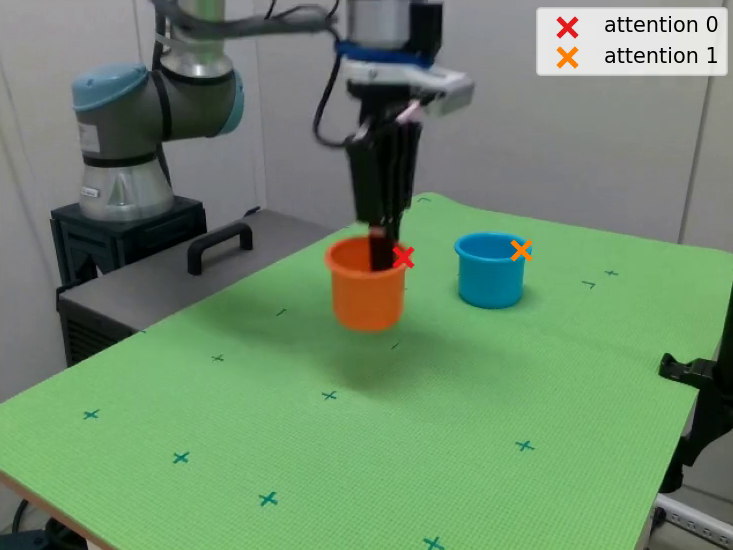}
  \includegraphics[width=0.39\columnwidth]{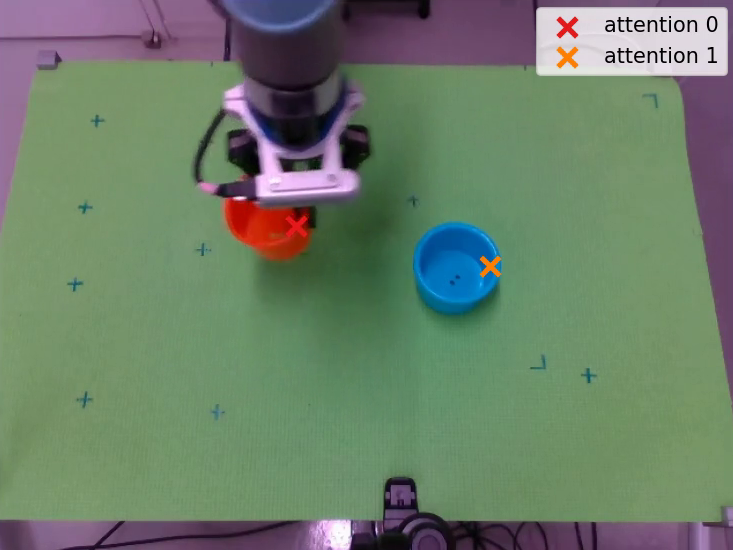}
  \includegraphics[width=0.39\columnwidth]{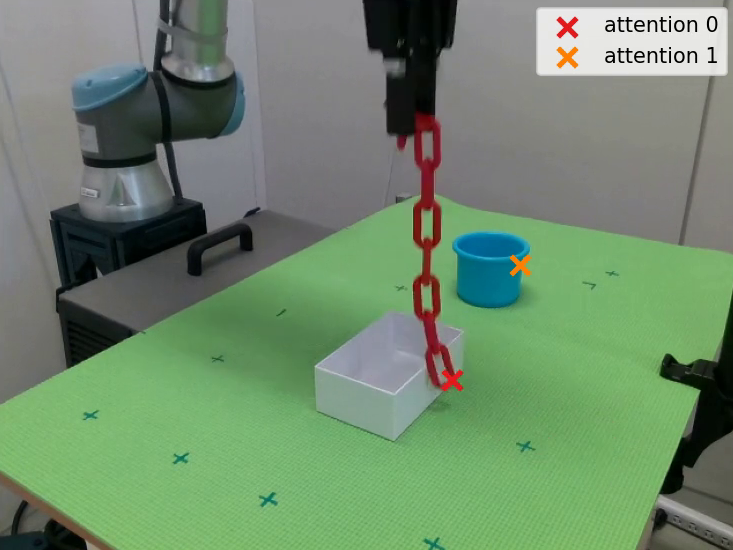}
  \includegraphics[width=0.39\columnwidth]{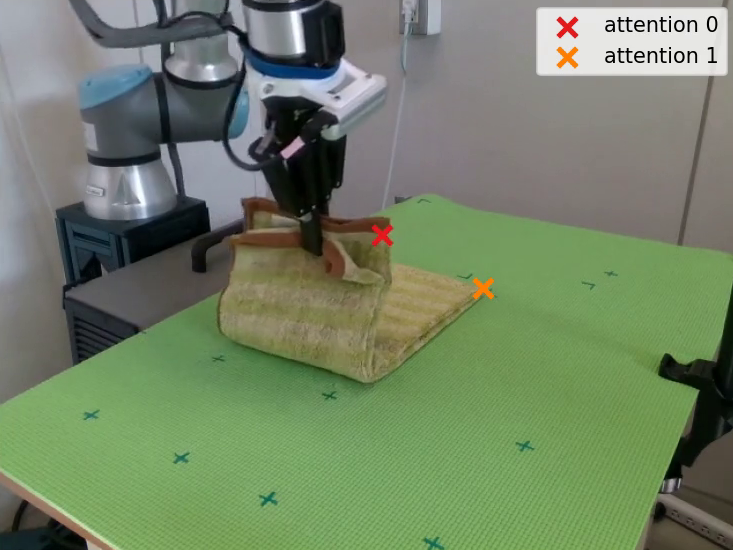}
  \includegraphics[width=0.39\columnwidth]{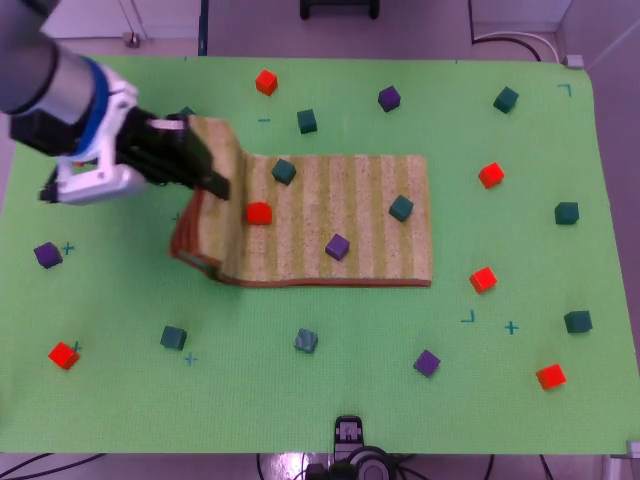}\\
  \vspace{1mm}
  \begin{minipage}{0.39\columnwidth}
    \begin{center} \footnotesize (A) Cup Insertion \end{center}
  \end{minipage}
  \begin{minipage}{0.39\columnwidth}
    \begin{center} \footnotesize (B) Cup Insertion (top view)  \end{center}
  \end{minipage}
  \begin{minipage}{0.39\columnwidth}
    \begin{center} \footnotesize (C) Chain Pick-and-Place \end{center}
  \end{minipage}
  \begin{minipage}{0.39\columnwidth}
    \begin{center} \footnotesize (D) Towel Fold \end{center}
  \end{minipage}
  \begin{minipage}{0.39\columnwidth}
    \begin{center} \footnotesize (E) Towel Fold \\(Appearance OOD) \end{center}
  \end{minipage}
  \caption{
    Real-world manipulation tasks.
    \newline
    \footnotesize{
      (A--D) show real-world task execution under ID conditions.
      (E) shows the Appearance OOD condition for Towel Fold,
      where numerous colorful block distractors introduce significant visual clutter.
      All tasks use two attention keypoints.
  }}
  \label{fig:real-tasks}
\end{figure*}

All real-world experiments are performed using a
Universal Robots UR5e manipulator with a parallel gripper.
Two RGB cameras (Intel RealSense D435) are used:
one mounted overhead and one mounted diagonally from the side.
The policy uses only the RGB streams as visual inputs.

We evaluate three manipulation tasks requiring visual precision
and generalization (\figref{fig:real-tasks}):

\textbf{Cup Insertion}:
The robot grasps an orange cup and inserts it into a slightly larger blue cup.
Two attention keypoints are used: the rim of each cup.

\textbf{Chain Pick-and-Place}:
The robot grasps a plastic chain and places it into a cup.
Two attention keypoints are used: the grasped chain tip and the cup rim.

\textbf{Towel Fold}:
The robot folds a rectangular towel in half on the tabletop.
Two attention keypoints are used: the two far corners of the towel.

For each task, we collect 32 teleoperated demonstrations:
8 demonstrations at each of 4 target-object positions.
Ground-truth attention keypoints are manually annotated
at the first frame and automatically propagated using Co-Tracker,
following the same protocol as in simulation.

\subsubsection{Evaluation Conditions}
Each method is evaluated over 3 random seeds, with 10 rollouts per condition.
We measure performance under ID and positional OOD conditions for all tasks, where ID rollouts are executed at target-object positions within the convex hull of the demonstrated target positions, and positional OOD rollouts are executed at previously unseen object locations outside that convex hull.
For the Towel Fold task, we additionally evaluate appearance OOD generalization by placing many small colorful blocks across the tabletop to induce strong texture changes.

\subsubsection{Quantitative Evaluation}

\begin{figure}[t]
  \centering
  \includegraphics[width=\columnwidth]{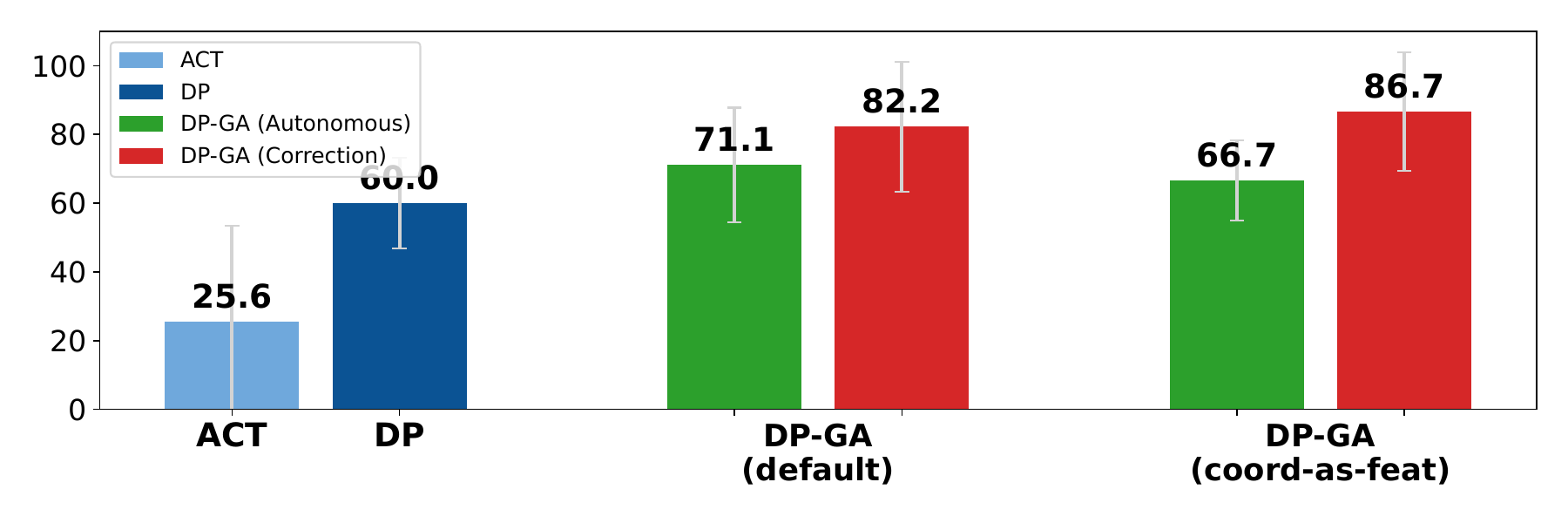}\\
  \vspace{-2mm}
  \footnotesize (A) ID performance\\
  \vspace{1mm}
  \includegraphics[width=\columnwidth]{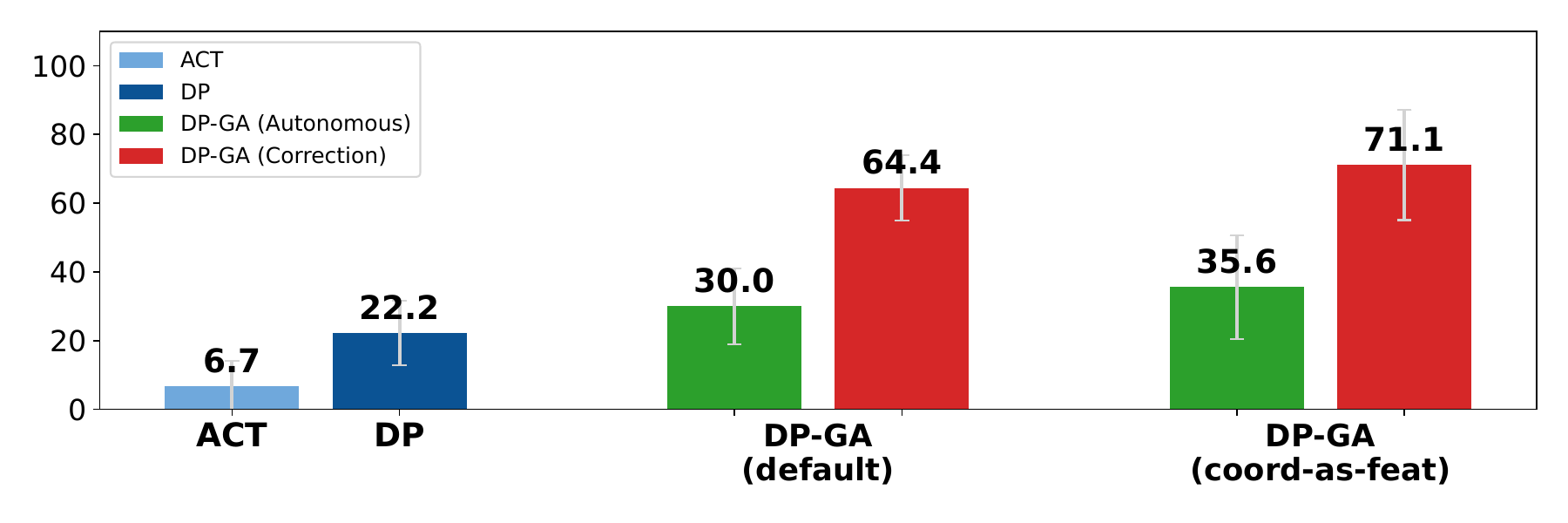}\\
  \vspace{-2mm}
  \footnotesize (B) Positional OOD performance\\
  \vspace{1mm}
  \includegraphics[width=\columnwidth]{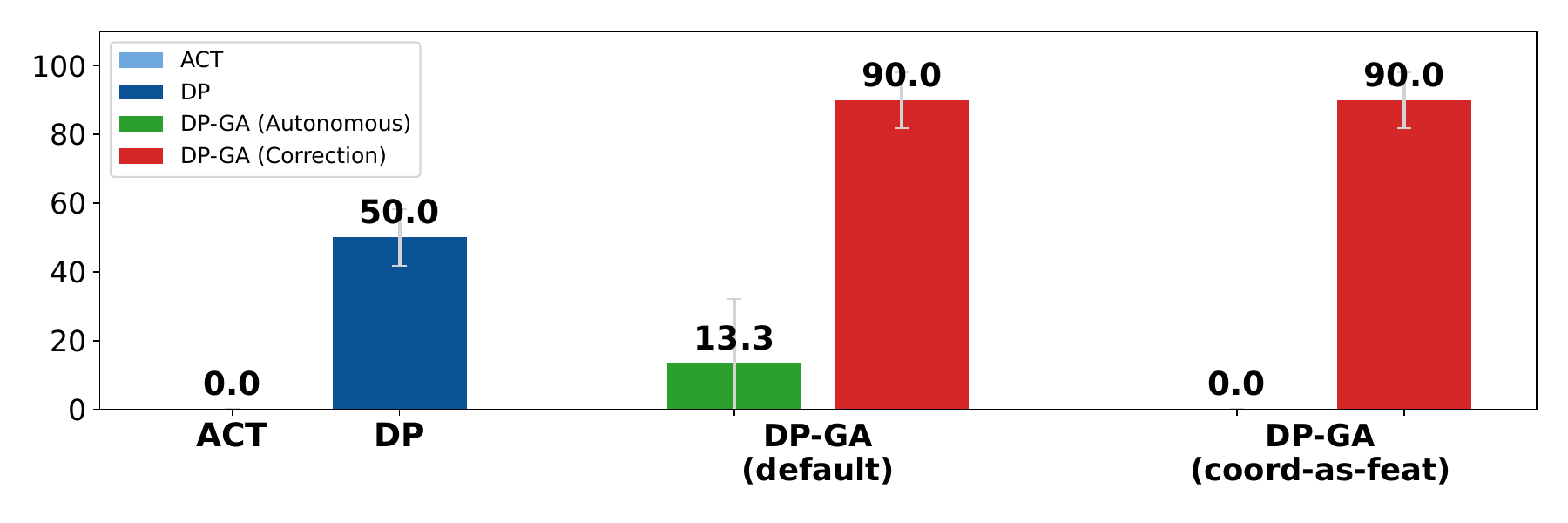}\\
  \vspace{-2mm}
  \footnotesize (C) Appearance OOD performance\\
  \vspace{1mm}
  \caption{Real-world manipulation success rates.
    \newline
    \footnotesize{
      Results in (A) and (B) are averaged across three real-world tasks.
      (C) is evaluated only on the Towel Fold task.
      All results are averaged over three random seeds with
      10 rollouts per condition for each seed.
      Blue bars indicate baseline policies.
      Green bars indicate fully autonomous execution.
      Red bars indicate optional human correction of the predicted attention.
  }}
  \label{fig:real-results}
\end{figure}

We evaluate the same four policies as in the simulation experiments:
DP-GA (default), DP-GA (coordinate-as-feature), DP baseline, and ACT baseline.
Each policy is evaluated both with and without attention correction.

\figref{fig:real-results} summarizes success rates averaged across the three
real-world tasks. Under ID conditions (\figref{fig:real-results}~(A)), both
variants of DP-GA outperform the baselines, confirming that leveraging
task-aligned attention cues improves visuomotor control even without
distribution shifts. DP-GA (default) achieves +15--30\,pp improvement
over ACT and +5--15\,pp improvement over DP.

Under positional OOD conditions (\figref{fig:real-results}~(B)), the advantage
of guided attention becomes more evident: even without attention correction,
DP-GA achieves slightly higher success rates than both baselines during fully
autonomous execution.
When optional human attention correction is provided, it further boosts performance,
yielding an additional +20--40\,pp improvement depending on the task.
In contrast, the baselines suffer substantial degradation, often failing to
complete the task.

Finally, appearance OOD is evaluated only on the Towel Fold task
(\figref{fig:real-results}~(C)). The severe visual distractors significantly
degrade performance for most methods, particularly those relying purely on
learned attention. While DP maintains a moderate level of robustness under
this appearance shift, human attention correction enables substantial recovery
in DP-GA, yielding roughly +80\,pp improvement over its autonomous execution
and about +40\,pp over the DP baseline.
These results show that optional correction of the predicted attention can
effectively mitigate perception failures while preserving autonomous policy
execution, even in highly challenging visual conditions.

\subsubsection{Qualitative Evaluation}

\begin{figure}[tpb]
  \centering
  \includegraphics[width=0.49\columnwidth]{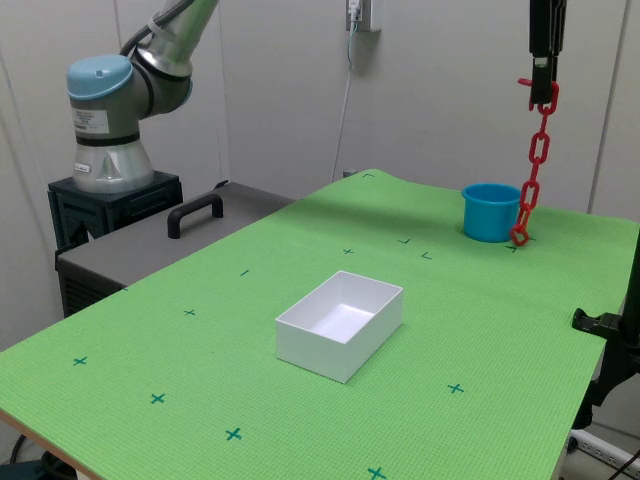}
  \includegraphics[width=0.49\columnwidth]{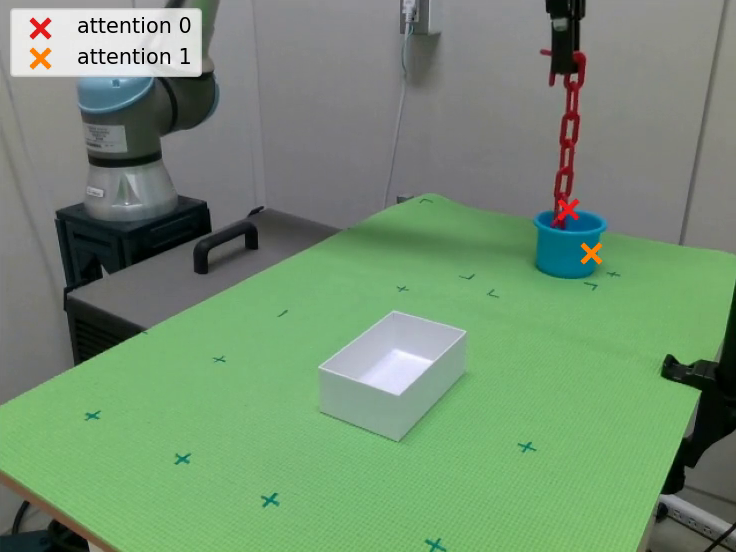}\\
  \vspace{1.5mm}
  \begin{minipage}{0.49\columnwidth}
    \begin{center} \footnotesize (A) DP baseline in Positional OOD (Failure) \end{center}
  \end{minipage}
  \begin{minipage}{0.49\columnwidth}
    \begin{center} \footnotesize (B) DP-GA (Autonomous) in Positional OOD (Success) \end{center}
  \end{minipage}\\
  \vspace{2mm}
  \includegraphics[width=0.49\columnwidth]{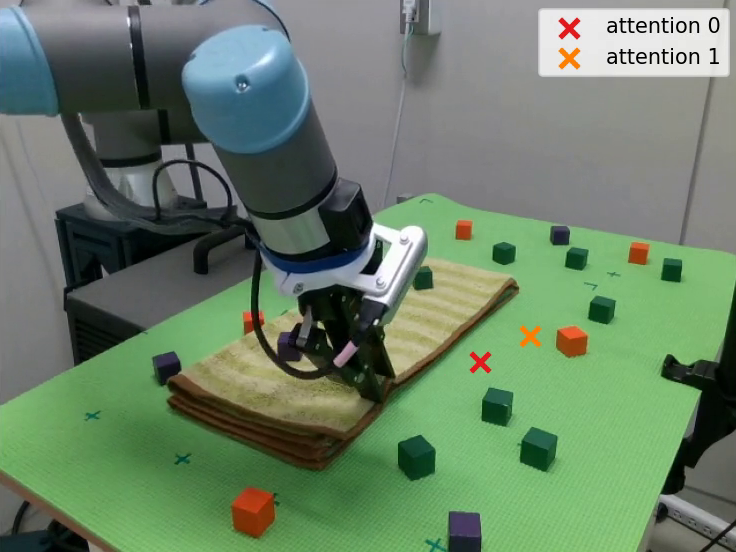}
  \includegraphics[width=0.49\columnwidth]{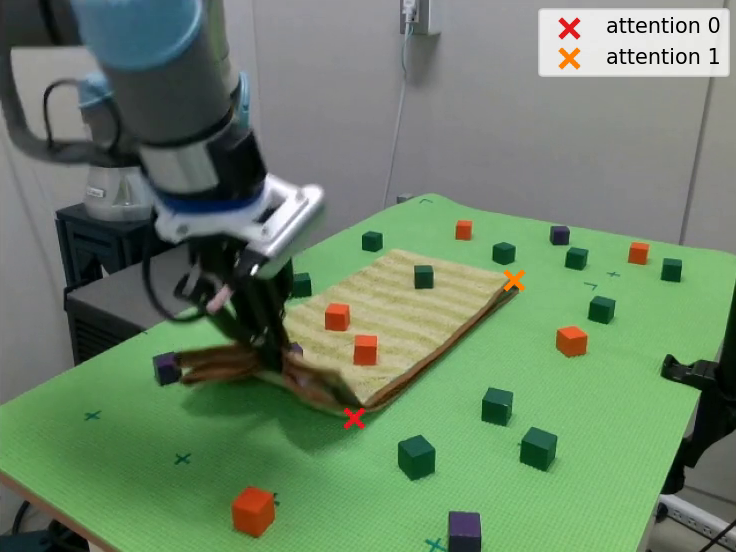}\\
  \vspace{1.5mm}
  \begin{minipage}{0.49\columnwidth}
    \begin{center} \footnotesize (C) DP-GA (Autonomous) in Appearance OOD (Failure) \end{center}
  \end{minipage}
  \begin{minipage}{0.49\columnwidth}
    \begin{center} \footnotesize (D) DP-GA (Attention Correction) in Appearance OOD (Success) \end{center}
  \end{minipage}
  \caption{
    Qualitative real-world rollouts under distribution shift.
    \newline
    \footnotesize{
      Guided visual attention improves robustness by maintaining focus on task-relevant
      structures.
      The predicted keypoints are visualized as \textbf{x}-marks on the images in DP-GA.\\
      (A, B) In the Chain Pick-and-Place task, DP-GA autonomously keeps
      attention near the chain end and around the cup rim, enabling successful
      placement even at unseen object positions.\\
      (C, D) In the Towel Fold task,
      accurate keypoints are provided only at the initial frame and then
      continuously used through automatic tracking, ensuring reliable grasping
      of the towel corner and successful folding under significant appearance
      changes.
  }}
  \label{fig:real-rollout}
\end{figure}

To better understand how visual attention contributes to task success or failure,
we visualize attention keypoints overlaid on camera images during real-world rollouts.

Figs.~\ref{fig:real-rollout}~(A) and (B) illustrate rollouts of the Chain
Pick-and-Place task under Positional OOD conditions using the DP baseline and
DP-GA (default), respectively.
Faced with previously unseen cup positions, the DP baseline fails to guide the
chain into the cup.
In contrast, DP-GA successfully completes the task by autonomously maintaining
relevant attention on the chain end and near the rim of the cup, despite the
positional shift.

Figs.~\ref{fig:real-rollout}~(C) and (D) compare Towel Fold rollouts under
Appearance OOD conditions, executed without and with attention correction.
Without attention correction, strong visual distractions cause incorrect
keypoint localization on the towel corner, resulting in failed grasp attempts.
In contrast, when the predicted attention is corrected only at the first frame
and then automatically tracked thereafter, the robot can reliably grasp the
towel corner and successfully complete the fold.

These qualitative results illustrate how interpretable and
correctable visual attention enhances spatial grounding and enables
robust visuomotor performance under real-world distribution shifts.

\section{Limitation} \label{sec:limitation}

While guided attention improves visuomotor robustness under environmental OOD, several limitations remain.

First, the approach assumes that a small number of manually defined keypoints is sufficient to represent all task-relevant structures. Although this assumption holds for the evaluated tasks, more complex scenarios may require richer or adaptive attention representations.

Second, keypoint annotations rely on 2D image coordinates and do not explicitly account for depth or occlusion, which may lead to degraded performance when keypoints are not visible.

Third, our method relies on an off-the-shelf tracking module to propagate keypoints over time, and its performance is therefore partly dependent on tracking quality. While tracking is generally stable in our settings, failures may occur in scenarios with severe deformation or prolonged occlusion. This limitation is partially alleviated by treating keypoints on static structures as fixed, but improving tracking robustness remains an important direction.

We believe addressing these points will further improve the applicability and robustness of guided attention.






\section{Conclusion} \label{sec:conclusion}

We presented GuidedAttention, a visuomotor imitation learning framework that
introduces interpretable and correctable visual attention as an explicit
intermediate representation between perception and action generation.
By allowing users to inspect and optionally correct the predicted attention
keypoints while preserving autonomous policy execution, GuidedAttention
provides an intuitive interface for human interaction with end-to-end
visuomotor policies.
Experimental results in both simulation and the real world demonstrate that
this human-correctable attention consistently improves manipulation
performance, particularly under positional and appearance distribution shifts.
We hope this work encourages future research on human-interpretable
representations and human-in-the-loop interaction for end-to-end robot
learning.

\bibliographystyle{IEEEtran}
\bibliography{main.bib}

\newpage

\setcounter{section}{0}
\renewcommand{\thesection}{A}
\renewcommand{\thefigure}{A\arabic{figure}}
\renewcommand{\thetable}{A\arabic{table}}
\setcounter{figure}{0}
\setcounter{table}{0}

\section*{APPENDIX}

\subsection{Comparison of Alternative Visual Prompts}

\subsubsection{Discussion of Prompt Representation Choices}

In imitation learning, several prior studies have explored the idea of allowing
users to provide visual prompts on camera images during policy rollout, similar
in spirit to our setting~\cite{RTTrajectory:Gu:ICLR2024,RoboticVisualInst:Li:CVPR2025,VPrompt:Muttaqien:CASE2025,RTSketch:Sundaresan:CoRL2024}.
However, these approaches differ from our method in several important respects.

First, they differ in the granularity of information required from the user.
In RT-Trajectory~\cite{RTTrajectory:Gu:ICLR2024}, the user specifies a continuous end-effector trajectory together with a grasp point.
While this allows detailed guidance to the policy, it imposes a substantially higher user burden than our method, which requires specifying only keypoints in the initial frame.
RT-Sketch~\cite{RTSketch:Sundaresan:CoRL2024} uses a line sketch of the goal
scene state as the prompt.
Although this input format is accessible even to users without robotics
expertise, it still requires drawing a sketch each time a manipulation command
is issued, which is not a negligible burden for the user.
Robotic Visual Instruction (RoVI)~\cite{RoboticVisualInst:Li:CVPR2025} allows
the user to write semantic instructions directly on the image.
It supports a variety of forms, such as object trajectories, numbers indicating execution order, and circles highlighting target objects.
Thus, in RoVI, the visual prompt represents the task intent itself, whereas in our method it serves only as a cue for the attention points maintained internally by the policy.
This design allows our method to position visual prompting strictly as an
auxiliary signal for out-of-distribution (OOD) situations.
As shown in Sec.~IV, even without manually overriding the attention keypoints,
our method still achieves higher success rates than baseline policies without
visual prompting, while the override further improves performance.


\subsubsection{Quantitative Comparison with Marker Overlays}

\begin{table*}[t]
\centering
\caption{Comparison with marker overlay prompting}
\label{tab:appendix-vp}
\begin{tabular}{lllll}
\toprule
Policy & ID & Pos-OOD & App-OOD & Pos+App-OOD \\
\midrule
DP-GA & \textbf{86.7\% {\tiny ($\pm$4.7\%)}} & \textbf{86.7\% {\tiny ($\pm$9.4\%)}} & \textbf{93.3\% {\tiny ($\pm$4.7\%)}} & \textbf{83.3\% {\tiny ($\pm$12.5\%)}} \\
MOP-Box & 60.0\% {\tiny ($\pm$0.0\%)} & 46.7\% {\tiny ($\pm$4.7\%)} & 43.3\% {\tiny ($\pm$26.2\%)} & 33.3\% {\tiny ($\pm$24.9\%)} \\
MOP-Point & 50.0\% {\tiny ($\pm$14.1\%)} & 50.0\% {\tiny ($\pm$8.2\%)} & 23.3\% {\tiny ($\pm$4.7\%)} & 3.3\% {\tiny ($\pm$4.7\%)} \\
\bottomrule
\end{tabular}\\
\vspace{2mm}
\begin{minipage}{1.8\columnwidth}
\footnotesize{
Comparison between DP-GA and Marker Overlay Prompting (MOP)~\cite{VPrompt:Muttaqien:CASE2025} under the same experimental conditions.
Results are shown for the Ring task in simulation.
Pos-OOD and App-OOD refer to OOD settings in object positions and table textures, respectively, that are not present in the training data.
Each value indicates mean success rate $\pm$ standard deviation across 3 random seeds,
with 10 rollouts for each ID / OOD condition.
Bold numbers denote the best performance in each condition.
}
\end{minipage}
\end{table*}

\begin{figure}[tpb]
  \centering
  \includegraphics[width=0.48\columnwidth]{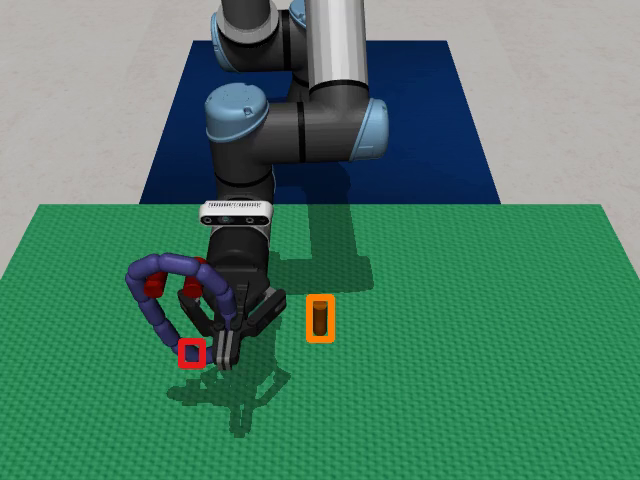}
  \includegraphics[width=0.48\columnwidth]{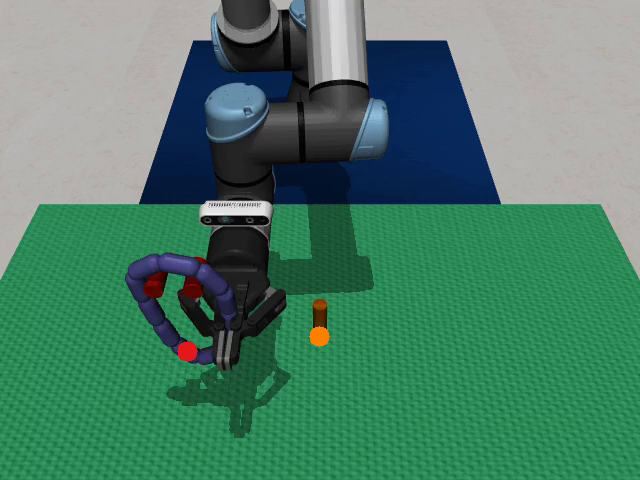}\\
  \vspace{0mm}
  \begin{minipage}{0.48\columnwidth}
    \begin{center} \footnotesize MOP-Box \end{center}
  \end{minipage}
  \begin{minipage}{0.48\columnwidth}
    \begin{center} \footnotesize MOP-Point \end{center}
  \end{minipage}
  \caption{Illustration of marker overlay prompting.}
  \label{fig:appendix-vp}
\end{figure}

As a baseline with relatively similar prompt granularity and purpose, we
compared our method with Marker Overlay Prompting (MOP), which overlays
fixed-color markers directly on the images provided to the policy during both
training and rollout~\cite{VPrompt:Muttaqien:CASE2025}.
To ensure a fair comparison, we matched the task and prompt locations, varying only the visual prompt representation.
We conducted this comparison in simulation on the Ring task shown in \figref{fig:sim-tasks}~(B), which involves placing a flexible ring around a pole.
As illustrated in \figref{fig:appendix-vp}, markers were overlaid at the same locations as the attention keypoints used in our method, specifically at the lowest point of the ring (red) and at the goal pole (orange).
In addition to MOP-Box, which overlays rectangular regions as in the prior method, we also evaluated MOP-Point, a variant that uses point-based overlays to better align with our representation.
As reported in Table~\ref{tab:appendix-vp}, our method consistently outperformed both baselines by more than 20 percentage points (pp) in success rate under both in-distribution (ID) and OOD conditions.
These results indicate that simply overlaying markers on the input image is insufficient for reliably exploiting task-relevant information. In contrast, our approach, which explicitly predicts attention points and leverages auxiliary supervision, is more effective in guiding the policy.


\subsection{Impact of the Tracking Module}

\subsubsection{Discussion on the Use of the Tracking Module}

Our method employs Co-Tracker~\cite{CoTracker:Karaev:ECCV2024}, an off-the-shelf tracking module, to automatically propagate attention keypoints over time. As a result, part of the overall performance depends on the quality of the tracker.
In the six simulation and real-world manipulation tasks considered in this paper, Co-Tracker was generally able to maintain stable tracking even when the target became temporarily occluded or moved outside the camera view and later reappeared. In these settings, we did not observe the tracker to be a dominant performance bottleneck.
However, in other task scenarios involving more severe deformation, tracking failures were observed, which may limit the applicability of the proposed method. This issue is partially mitigated in our framework, as keypoints attached to fixed structures in the environment (e.g., a pole mounted on a table) are defined as static keypoints and are therefore not affected by the tracking module.
In addition, the method can be readily integrated with improved tracking modules, should such advances become available in the future.


\subsubsection{Quantitative Impact of Tracking Accuracy}

\begin{figure}[tpb]
  \centering
  \includegraphics[width=0.98\columnwidth]{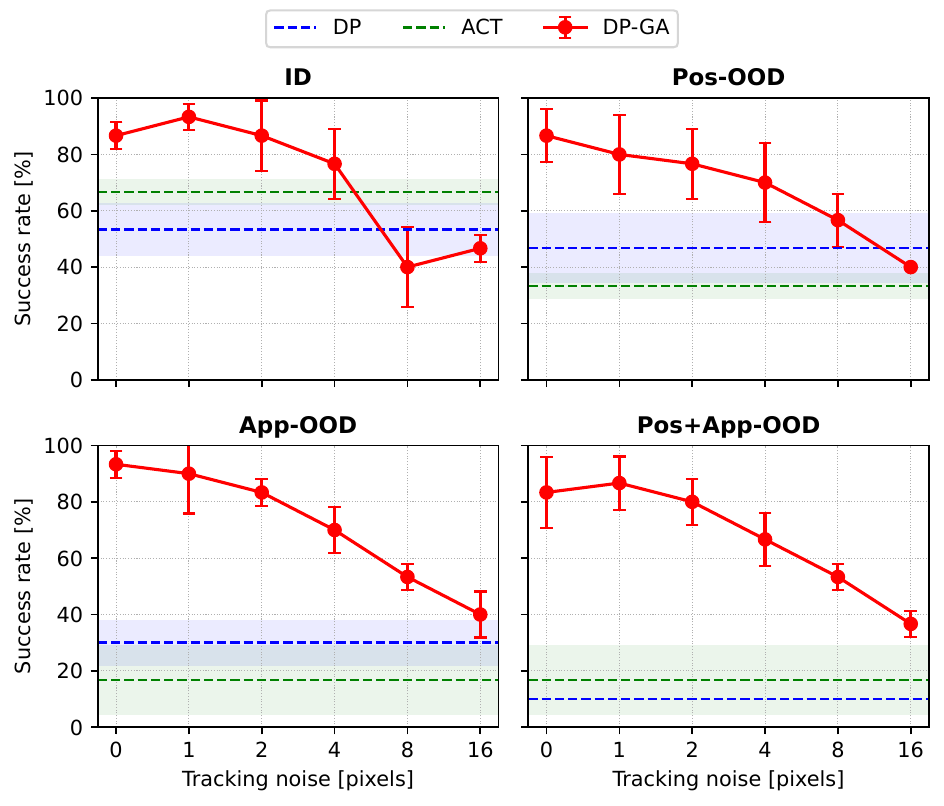}
  \caption{Quantitative impact of tracking accuracy.
    \newline
    \footnotesize{
      Success rates on the Ring task in simulation for three different policies, evaluated across three random seeds while varying the magnitude of noise added to the tracked keypoints.
      Pos-OOD and App-OOD refer to OOD settings in object positions and table textures, respectively.
  }}
  \label{fig:appendix-tracker}
\end{figure}

We conducted an evaluation in simulation on the Ring task shown in \figref{fig:sim-tasks}~(B) by injecting random noise of varying magnitude into the keypoints tracked by Co-Tracker during both training and rollout, and measuring the resulting success rates.
As shown in \figref{fig:appendix-tracker}, the success rate remains largely stable within a moderate noise range, and gradually decreases as the noise magnitude increases.
Under a setting with shifts in both object positions and background appearance (Pos+App-OOD), with an input resolution of $320 \times 224$, when the noise magnitude is 8 pixels, the success rate decreases by approximately 30 pp; however, it still remains more than 20 pp higher than those of baselines without attention keypoints.


\subsection{Impact of Keypoint Annotation Accuracy}

\begin{figure}[tpb]
  \centering
  \includegraphics[width=0.98\columnwidth]{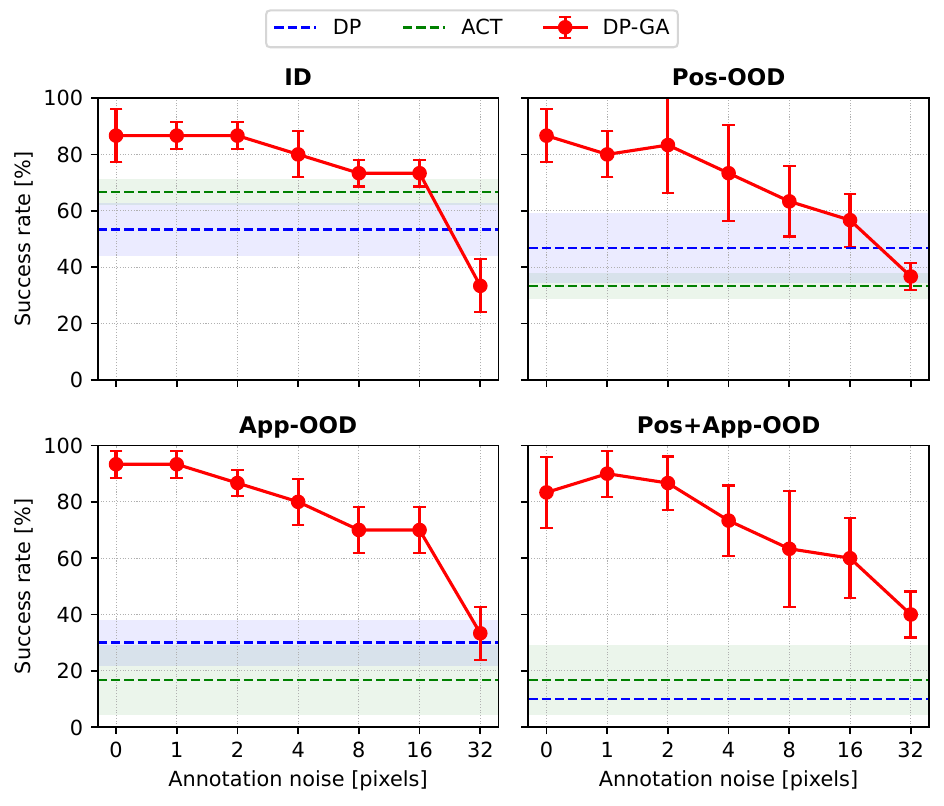}
  \caption{Quantitative impact of keypoint annotation accuracy.
    \newline
    \footnotesize{
      Success rates on the Ring task in simulation for three different policies, evaluated across three random seeds while varying the magnitude of noise added to the annotated keypoints.
      Pos-OOD and App-OOD refer to OOD settings in object positions and table textures, respectively.
  }}
  \label{fig:appendix-annotation}
\end{figure}

User-specified attention keypoints are annotated by clicking points on the camera image, and such inputs can naturally contain annotation errors.
To quantitatively evaluate the tolerance of the proposed method to annotation errors, we conducted a simulation evaluation on the Ring task.
At rollout time, we injected random noise of varying magnitude into the attention keypoints specified by the user in the initial frame and evaluated the resulting success rates.
The results are shown in \figref{fig:appendix-annotation}. While the success rate is largely preserved under small perturbations, it gradually decreases as the noise magnitude increases.
Under a setting with shifts in both object positions and background appearance (Pos+App-OOD), with an input resolution of $320 \times 224$, when the noise magnitude is 8 pixels, the success rate decreases by approximately 20 pp; however, it still remains more than 40 pp higher than those of baselines without attention keypoints.
In our experience, this level of sensitivity is sufficiently mild that users can provide annotations without difficulty while still benefiting from the performance of the proposed method.

\subsection{Impact of the Number of Keypoints}

In the proposed method, users are not required to override all attention keypoints during rollout, and can instead specify only a subset. To evaluate the impact of the number of overridden keypoints, we conducted a simulation study on the Ring task.
Specifically, we trained policies with 2, 4, and 6 attention keypoints, and for each policy, varied the number of keypoints overridden at rollout while measuring task success rates. The results are shown in Table~\ref{tab:appendix-num}.
Overall, performance tends to improve as a larger proportion of keypoints is overridden at rollout. However, increasing the total number of keypoints used during training does not significantly change the success rate.
Keypoints are mainly placed around the ring and the pole, and overriding those near the pole tends to yield higher success rates.
This indicates that the importance of keypoints depends on the associated object, with some keypoints having a stronger impact on task success than others.
When task-relevant keypoints are selected appropriately, high success rates can be achieved with as few as two keypoints. This allows users to benefit from the proposed method without a substantial increase in annotation effort.


\begin{table*}[t]
\centering
\caption{Quantitative impact of the number of keypoints}
\label{tab:appendix-num}
\setlength{\tabcolsep}{4pt}
\renewcommand{\arraystretch}{1.1}
\vspace{-6mm}
\begin{tabular}{cc}
\begin{tabular}{c}
\vspace{0.5em}\\
\small \textbf{ID} \\
\begin{tabular}{c|ccccc}
\toprule
 & \multicolumn{5}{c}{$N_k$ (override at rollout)} \\
$N_k$ (train) & 1 (ring) & 1 (pole) & 2 & 4 & 6 \\
\midrule
2 & 76.7\% & 76.7\% & 86.7\% & --   & -- \\
4 &      &      & 86.7\% & 96.7\% & -- \\
6 &      &      & 80.0\% & 93.3\% & 93.3\% \\
\bottomrule
\end{tabular}
\end{tabular}
&
\begin{tabular}{c}
\vspace{0.5em}\\
\small \textbf{Pos-OOD} \\
\begin{tabular}{c|ccccc}
\toprule
 & \multicolumn{5}{c}{$N_k$ (override at rollout)} \\
$N_k$ (train) & 1 (ring) & 1 (pole) & 2 & 4 & 6 \\
\midrule
2 & 50.0\% & 80.0\% & 86.7\% & --   & -- \\
4 &      &      & 73.3\% & 93.3\% & -- \\
6 &      &      & 70.0\% & 80.0\% & 93.3\% \\
\bottomrule
\end{tabular}
\end{tabular}
\\[0.0em]
\begin{tabular}{c}
\vspace{0.5em}\\
\small \textbf{App-OOD} \\
\begin{tabular}{c|ccccc}
\toprule
 & \multicolumn{5}{c}{$N_k$ (override at rollout)} \\
$N_k$ (train) & 1 (ring) & 1 (pole) & 2 & 4 & 6 \\
\midrule
2 & 30.0\% & 80.0\% & 93.3\% & --   & -- \\
4 &      &      & 43.3\% & 96.7\% & -- \\
6 &      &      & 23.3\% & 46.7\% & 96.7\% \\
\bottomrule
\end{tabular}
\end{tabular}
&
\begin{tabular}{c}
\vspace{0.5em}\\
\small \textbf{Pos+App-OOD} \\
\begin{tabular}{c|ccccc}
\toprule
 & \multicolumn{5}{c}{$N_k$ (override at rollout)} \\
$N_k$ (train) & 1 (ring) & 1 (pole) & 2 & 4 & 6 \\
\midrule
2 & 16.7\% & 56.7\% & 83.3\% & --   & -- \\
4 &      &      & 46.7\% & 96.7\% & -- \\
6 &      &      & 16.7\% & 50.0\% & 80.0\% \\
\bottomrule
\end{tabular}
\end{tabular}
\end{tabular}\\
\vspace{4mm}
\begin{minipage}{1.8\columnwidth}
\footnotesize{
Results on the Ring task in simulation are shown.
Pos-OOD and App-OOD refer to OOD settings in object positions and table textures, respectively.
Each value indicates the mean success rate over 3 random seeds, with 10 rollouts per ID/OOD condition.
}
\end{minipage}
\vspace{2mm}
\end{table*}

\subsection{Impact of the Number of Cameras}

\begin{table*}[t]
\centering
\caption{Quantitative impact of the number of cameras}
\label{tab:app-camera}
\setlength{\tabcolsep}{4pt}
\renewcommand{\arraystretch}{1.1}
\vspace{-6mm}
\begin{tabular}{cc}
\begin{tabular}{c}
\vspace{0.5em}\\
\small \textbf{ID} \\
\begin{tabular}{l|cc|c}
\toprule
\makecell[l]{Policy input\\images} & \makecell[c]{DP-GA\\(front-only override)} & \makecell[c]{DP-GA\\(both-image override)} & DP \\
\midrule
front & 86.7\% & - & 56.7\% \\
front+left & 80.0\% & 86.7\% & 56.7\% \\
front+right & 83.3\% & \textbf{93.3\%} & 50.0\% \\
\bottomrule
\end{tabular}
\end{tabular}
&
\begin{tabular}{c}
\vspace{0.5em}\\
\small \textbf{Pos-OOD} \\
\begin{tabular}{l|cc|c}
\toprule
\makecell[l]{Policy input\\images} & \makecell[c]{DP-GA\\(front-only override)} & \makecell[c]{DP-GA\\(both-image override)} & DP \\
\midrule
front & \textbf{86.7\%} & - & 60.0\% \\
front+left & 60.0\% & 63.3\% & 50.0\% \\
front+right & 80.0\% & \textbf{83.3\%} & 43.3\% \\
\bottomrule
\end{tabular}
\end{tabular}
\\[0.0em]
\begin{tabular}{c}
\vspace{0.5em}\\
\small \textbf{App-OOD} \\
\begin{tabular}{l|cc|c}
\toprule
\makecell[l]{Policy input\\images} & \makecell[c]{DP-GA\\(front-only override)} & \makecell[c]{DP-GA\\(both-image override)} & DP \\
\midrule
front & \textbf{93.3\%} & - & 23.3\% \\
front+left & 30.0\% & 50.0\% & 26.7\% \\
front+right & 83.3\% & \textbf{90.0\%} & 53.3\% \\
\bottomrule
\end{tabular}
\end{tabular}
&
\begin{tabular}{c}
\vspace{0.5em}\\
\small \textbf{Pos+App-OOD} \\
\begin{tabular}{l|cc|c}
\toprule
\makecell[l]{Policy input\\images} & \makecell[c]{DP-GA\\(front-only override)} & \makecell[c]{DP-GA\\(both-image override)} & DP \\
\midrule
front & \textbf{83.3\%} & - & 10.0\% \\
front+left & 33.3\% & 36.7\% & 23.3\% \\
front+right & 43.3\% & \textbf{80.0\%} & 26.7\% \\
\bottomrule
\end{tabular}
\end{tabular}
\end{tabular}\\
\vspace{4mm}
\begin{minipage}{1.8\columnwidth}
\footnotesize{
Results on the Ring task in simulation are shown.
Pos-OOD and App-OOD refer to OOD settings in object positions and table textures, respectively.
Bold numbers denote the best performance and those within 5 percentage points of the best in each condition.
Each value indicates the mean success rate over 3 random seeds, with 10 rollouts per ID/OOD condition.
}
\end{minipage}
\vspace{2mm}
\end{table*}

\begin{figure*}[tpb]
  \centering
  \includegraphics[width=0.45\columnwidth]{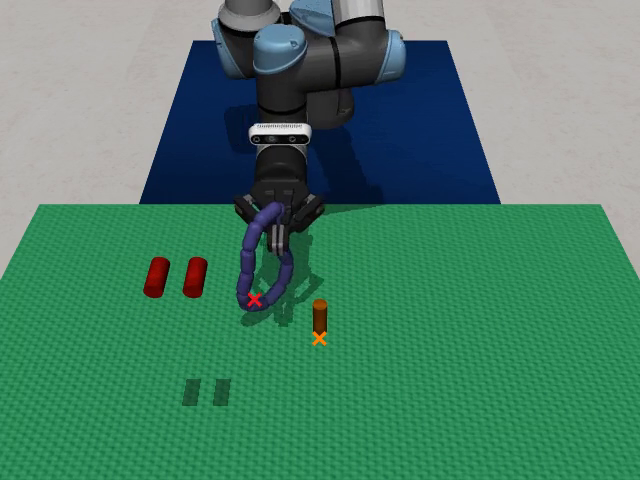}
  \includegraphics[width=0.45\columnwidth]{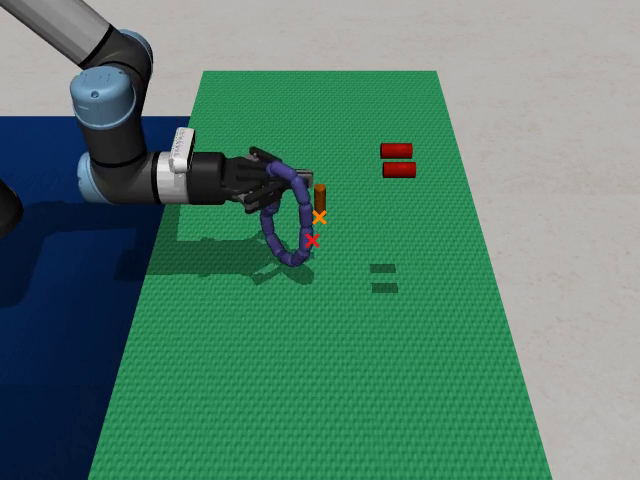}
  \includegraphics[width=0.45\columnwidth]{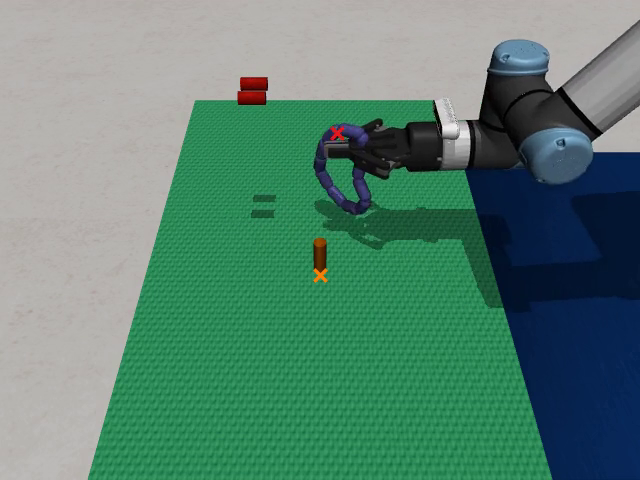}\\
  \vspace{1mm}
  \begin{minipage}{0.45\columnwidth}
    \begin{center} \footnotesize (A) Front camera \end{center}
  \end{minipage}
  \begin{minipage}{0.45\columnwidth}
    \begin{center} \footnotesize (B) Left camera  \end{center}
  \end{minipage}
  \begin{minipage}{0.45\columnwidth}
    \begin{center} \footnotesize (C) Right camera \end{center}
  \end{minipage}
  \vspace{-2mm}
  \caption{Camera images from multiple viewpoints.}
  \label{fig:app-camera}
\end{figure*}

As shown in \figref{fig:policy}~(A), the proposed policy architecture can take multiple camera images as input. Furthermore, attention keypoints can be overridden at rollout time for only a subset of the cameras.
To evaluate the impact of the number of cameras, we conducted a simulation study on the Ring task. In addition to the original front-view camera, we introduced left- and right-view cameras in the workspace, as shown in \figref{fig:app-camera}. We trained policies with two-camera inputs (front+left and front+right).
For the two-camera policies, we further evaluated two settings at rollout: overriding keypoints only for the front camera, and overriding keypoints for both cameras.
The results are shown in Table~\ref{tab:app-camera}. Our DP-GA consistently achieves higher success rates than the DP baseline without attention keypoints. Among the DP-GA variants, the front-only and front+right policies achieve similarly high performance, while the front+left policy shows lower success rates. This is likely because, from the left camera viewpoint, the ring and the pole are more frequently occluded, making it harder to distinguish task-relevant structures.
These results indicate that the proposed method can effectively leverage multiple cameras. However, increasing the number of cameras does not necessarily improve performance; rather, selecting viewpoints that provide task-relevant observations is more important.


\end{document}